\documentclass[sigconf, nonacm]{acmart}

\usepackage{graphicx}
\usepackage{listings}

\usepackage{enumitem}
\usepackage{color}
\usepackage{float}                  
\usepackage{subfig}                 
\usepackage{overpic}                
\definecolor{dkgreen}{rgb}{0,0.6,0}
\definecolor{gray}{rgb}{0.5,0.5,0.5}
\definecolor{mauve}{rgb}{0.58,0,0.82}
\definecolor{Myblue}{rgb}{0,0.5,0.8}

\usepackage{amsmath}

\usepackage{amssymb}
\newtheorem{theorem}{Theorem}
\newtheorem{definition}{Definition}

\usepackage[linesnumbered,ruled,vlined]{algorithm2e}
\usepackage{tabularx,booktabs}
\usepackage{makecell}
\usepackage{multirow}
\usepackage{multicol}
\usepackage{utfsym}

\newcommand\vldbdoi{10.14778/3632093.3632108}
\newcommand\vldbpages{455 - 468}
\newcommand\vldbvolume{17}
\newcommand\vldbissue{3}
\newcommand\vldbyear{2023}

\newcommand\vldbtitle{\shorttitle} 
\newcommand\vldbavailabilityurl{https://github.com/iDC-NEU/NeutronStream}
\newcommand\vldbpagestyle{empty} 

\newcommand{\wqg}[1]{\textcolor{blue}{#1}}

\newcommand{\oursys}{NeutronStream\xspace}
\newcommand{\Paragraph} [1] {\smallskip\noindent{\bf #1. }}
\newcommand{\eat}[1]{}

\newcommand{\addrevision}[1]{\textcolor{black}{#1}}

\newcommand{\ie}{\emph{i.e.,}\xspace}
\newcommand{\eg}{\emph{e.g.,}\xspace}

\newcommand{\batch}{\texttt{NS-Batch}\xspace}
\newcommand{\sliding}{\texttt{NS-Slide}\xspace}
\newcommand{\torchbatch}{\texttt{Torch-Batch}\xspace}
\newcommand{\adasliding}{\texttt{NS-AdaSlide}\xspace}

\begin{document}
\title{NeutronStream: A Dynamic GNN Training Framework with Sliding Window for Graph Streams}

%





\vspace{-0.35in}
\settopmatter{authorsperrow=5}
\author{Chaoyi Chen}
\affiliation{%
  \institution{Northeastern University\country{China} \\ chenchaoy@\\stumail.neu.edu.cn}
}

\author{Dechao Gao}
\affiliation{%
  \institution{Northeastern University\country{China} \\ gaodechao@\\stumail.neu.edu.cn}
}

\author{Yanfeng Zhang}
\affiliation{%
  \institution{Northeastern University\country{China} \\ zhangyf@\\mail.neu.edu.cn}
}

\author{Qiange Wang}
\affiliation{%
  \institution{Northeastern University\country{China\\wangqiange@\\stumail.neu.edu.cn}}
}

\author{Zhenbo Fu}
\affiliation{%
  \institution{Northeastern University\country{China} \\ fuzhenbo@\\stumail.neu.edu.cn}
}

\author{Xuecang Zhang}
\affiliation{%
  \institution{Huawei Technologies Co., Ltd. \\ zhangxuecang@\\huawei.com}
}

\author{Junhua Zhu}
\affiliation{%
  \institution{Huawei Technologies Co., Ltd. \\ junhua.zhu@\\huawei.com}
}

\author{Yu Gu}
\affiliation{%
  \institution{Northeastern University\country{China} \\ guyu@\\mail.neu.edu.cn}
}

\author{Ge Yu}
\affiliation{%
  \institution{Northeastern University\country{China} \\ yuge@\\mail.neu.edu.cn}
}
\vspace{-0.15in}


\begin{abstract}

Existing Graph Neural Network (GNN) training frameworks have been designed to help developers easily create performant GNN implementations.
However, most existing GNN frameworks assume that the input graphs are static, but ignore that most real-world graphs are constantly evolving. Though many dynamic GNN models have emerged to learn from evolving graphs, the training process of these dynamic GNNs is dramatically different from traditional GNNs in that it captures both the spatial and temporal dependencies of graph updates. This poses new challenges for designing dynamic GNN training frameworks.
First, the traditional batched training method fails to capture real-time structural evolution information. Second, the time-dependent nature makes parallel training hard to design. Third, it lacks system supports for users to efficiently implement dynamic GNNs.
In this paper, we present \oursys, a framework for training dynamic GNN models. \oursys abstracts the input dynamic graph into a chronologically updated stream of events and processes the stream with an optimized sliding window to incrementally capture the spatial-temporal dependencies of events. Furthermore, \oursys provides a parallel execution engine to tackle the sequential event processing challenge to achieve high performance. \oursys also integrates a built-in graph storage structure that supports dynamic updates and provides a set of easy-to-use APIs that allow users to express their dynamic GNNs.
Our experimental results demonstrate that, compared to state-of-the-art dynamic GNN implementations, \oursys achieves speedups ranging from 1.48X to 5.87X and an average accuracy improvement of 3.97\%.
\end{abstract}

\maketitle


\vspace{-0.13in}

\pagestyle{\vldbpagestyle}
\begingroup\small\noindent\raggedright\textbf{PVLDB Reference Format:}\\
{Chaoyi Chen, Dechao Gao, Yanfeng Zhang, Qiange Wang, Zhenbo Fu,
Xuecang Zhang, Junhua Zhu, Yu Gu, Ge Yu}. \vldbtitle. PVLDB, \vldbvolume(\vldbissue): \vldbpages, \vldbyear.\\
\href{https://doi.org/\vldbdoi}{doi:\vldbdoi}
\endgroup

\begingroup
\vspace{-0.18in}
\renewcommand\thefootnote{}\footnote{\noindent
This work is licensed under the Creative Commons BY-NC-ND 4.0 International License. Visit \url{https://creativecommons.org/licenses/by-nc-nd/4.0/} to view a copy of this license. For any use beyond those covered by this license, obtain permission by emailing \href{mailto:info@vldb.org}{info@vldb.org}. Copyright is held by the owner/author(s). Publication rights licensed to the VLDB Endowment. \\
\raggedright Proceedings of the VLDB Endowment, Vol. \vldbvolume, No. \vldbissue\ %
ISSN 2150-8097. \\
\href{https://doi.org/\vldbdoi}{doi:\vldbdoi} \\
}\addtocounter{footnote}{-1}\endgroup

\vspace{-0.11in}
\ifdefempty{\vldbavailabilityurl}{}{
\vspace{.3cm}
\begingroup\small\noindent\raggedright\textbf{PVLDB Artifact Availability:}\\
The source code, data, and/or other artifacts have been made available at \url{\vldbavailabilityurl}.
\endgroup
}

\vspace{-0.10in}
\section{Introduction}
\label{sec:introduction}

\begin{figure*}[t]
\vspace{-0.2in}
  \centering  \includegraphics[width=6.2in]{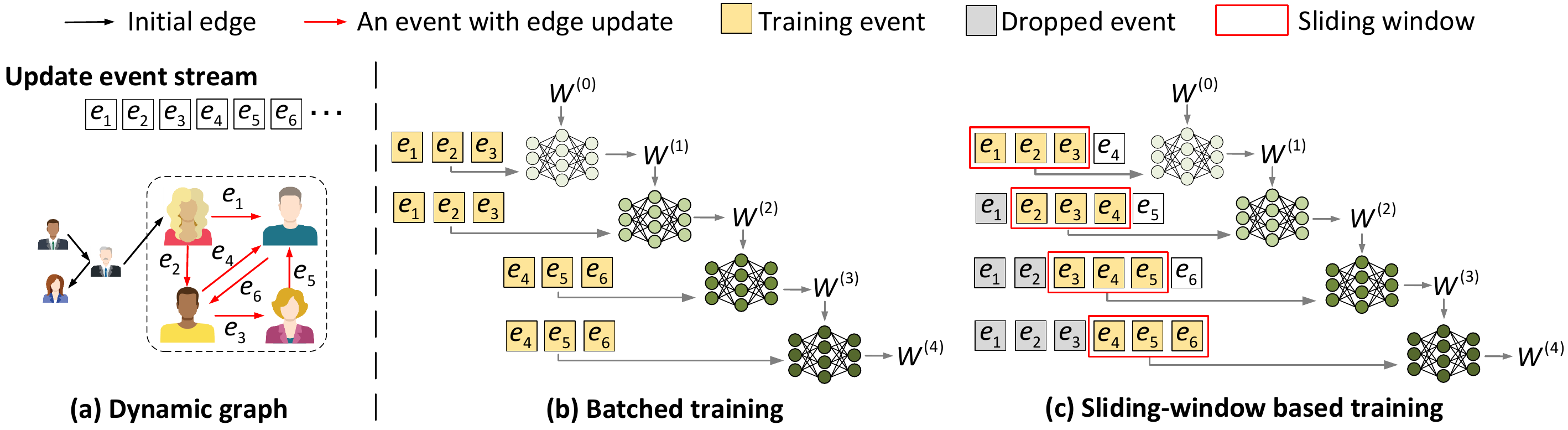}
  \vspace{-0.1in}
  \captionsetup{font=small}
  \caption{Traditional batched training vs. sliding-window based training for graph streams.}
  \label{fig:intro}
  \vspace{-0.10in}
\end{figure*}

Graph Neural Networks (GNNs) \cite{GGSNN_ICLR_2016, HGNN_CoRR_2019, CNNonGraph_NIPS_2016, ESGCN_EMNLP_2017, 
GCN_ICLR_2017, GIN_ICLR_2019, gnn_survey_arxiv2019, GAT_ICLR_2018, GNNRS_CoRR_2020} are a class of deep learning models designed to learn from graph data.
GNNs have been widely adopted in various graph applications, including social networks analytics \cite{DRKGFORSOCIAL_IJCAI_2018, GNNSocialRec_arxiv_2019}, recommendation systems \cite{GMCRMGNN_NIPS_2017, GCNFORWEBRREC_KDD_2018}, and knowledge graphs \cite{RGCN_ESWC_2018, ENIKG_SIGKDD_2019}. 
Most of the existing GNN models assume that the input graph is static. 
However, real-world graphs are inherently dynamic and evolving over time.  
Recently, many dynamic GNN models \cite{Hawkes_KDD_2018, JODIE_arxiv_2018, DyRep_ICLR_2019, MMDNE_CIKM_2019,  TGN_2020_arXiv, DyGNN_SIGIR_2020, TGAT_2020_arXiv, APAN_SIGMOD_2021, FDGNN_arxiv_2022} are emerged as a promising method for learning from dynamic graphs. These models capture both the \textit{spatial} and \textit{temporal} information, which makes them outperform traditional GNNs in real-time applications, such as real-time fraud detection \cite{APAN_SIGMOD_2021}, real-time recommendation \cite{JODIE_arxiv_2018}, and many other tasks.





In dynamic GNNs, the dynamic graph is modeled as a sequence of time-stamped \textit{events}, each event representing a graph update operation.
Each event is associated with a timestamp indicating when it occurs and an update type, \eg an addition/deletion of a node/edge or an update of a node/edge's feature. Dynamic GNNs encode the information of each event into dynamic node embeddings chronologically. The training process of dynamic GNNs is dramatically different from traditional GNNs in that it has to consider the temporal dependency of events. Existing dynamic GNNs \cite{DyRep_ICLR_2019, LDG_Plosone_2021, DyGNN_SIGIR_2020} are implemented on top of general DNN training frameworks, \eg Tensorflow \cite{Tensorflow_OSDI_2016} and PyTorch \cite {Pytorch_NIPS_2019}. However, the complex spatial-temporal dependencies among events pose new challenges for designing dynamic GNN frameworks.

First, the traditional batched training mode adopted by existing DNN frameworks may fail to capture the real-time structural evolution information. 
Batched training mode periodically packs new arrival events into a training batch and trains the model using these batches incrementally. However, this method forcibly cuts off the stream and ignores the spatial locality between events in two consecutive batches, which may lead to a decrease in model accuracy. Figure \ref{fig:intro}(a) illustrates a motivating example on a dynamic social network graph, which contains six consecutive interaction events concentrated in the rectangular area. In the batched training mode, as shown in Figure \ref{fig:intro}(b), these six events with spatial locality are split into two independent batches. With an initial parameter matrix $W^{(0)}$, the first batch with events $\{e_1, e_2, e_3\}$ is trained in two epochs, and the parameter matrix is updated twice to obtain $W^{(2)}$. The second batch with events $\{e_4, e_5, e_6\}$ is then trained in two epochs to obtain $W^{(4)}$. In this way, the spatial dependency between cross-batch events (\eg $e_3$ and $e_4$) cannot be captured, thereby impacting the model accuracy.

Second, the sequential nature of time-dependent events makes parallelism optimization hard to design. The existing open-source dynamic GNN implementations \cite{DyRep_ICLR_2019, LDG_Plosone_2021, DyGNN_SIGIR_2020} are based on the mature DNN or GNN frameworks (\eg Pytorch \cite{Pytorch_NIPS_2019} and DGL\cite{DGL_arXiv_2019}). They adopt vanilla sequential iterative processing, which suffers from sub-optimal performance and can hardly benefit from modern parallel architectures. For example, the DyRep \cite{DyRep_ICLR_2019} implementation in PyTorch takes 60.45 seconds with a single thread for an epoch on the GitHub dataset \cite{github_dataset}, but it takes 60.89 seconds with 20 threads, which is even longer than single-thread execution, indicating the lack of parallelization support.

Third, there lacks a user-friendly programming abstraction for implementing dynamic GNNs. The training of dynamic GNNs highly relies on efficient dynamic storage for maintaining dynamic graphs, which is not a standard module in traditional GNN frameworks. Users have to implement their own store to maintain graph topology and multi-versioned embeddings. For example, in the Dyrep \cite{DyRep_ICLR_2019} model, users have to implement data structures to support the storage of timestamps, the update of graph and node embeddings, and complex graph operations such as accessing a node’s neighbor nodes and their embeddings. This requires much redundant coding work and is error-prone. Therefore, it is desirable to have a set of specific APIs for implementing dynamic GNNs and an efficient framework that supports common modules used in dynamic GNN training.





To address the above problems for dynamic GNN training, we make the following contributions.

\Paragraph{Contribution 1} We propose a new incremental learning mode with a sliding window for training on graph streams. We rely on a sliding window to select consecutive events from the graph stream feeding to the model training. The window is sliding as new update events arrive and the processed stale events outside of the window can be dropped, so that the freshness of the model can be guaranteed and the evolving information and temporal dependencies can be well captured. For example, in Figure \ref{fig:intro}(c), a window of 3 events as training samples are trained for one epoch, and then the window slides with a sliding stride of 1. 
The window size can be adaptively adjusted for capturing the complete spatial dependencies.

\Paragraph{Contribution 2} We propose a fine-grained event parallel execution scheme.
We leverage the observation that each event only affects a small subgraph, and there is no read-write conflict between multiple events if their affected subgraphs do not intersect.
Based on this observation, we build a dependency graph analysis method that identifies the events having no node-updating conflicts and processes them in parallel. In this way, training performance can be enhanced through fine-grained event parallelism.

\Paragraph{Contribution 3} We deliver a dynamic GNN training framework \oursys. \oursys integrates sliding-window training method and dependency-graph-driven event parallelizing method. Moreover, \oursys integrates a built-in graph storage structure that supports dynamic updates and provides a set of easy-to-use APIs that allow users to express their dynamic GNNs with lightweight engineering work.

We evaluate \oursys on three popular dynamic GNN models, DyRep, LDG, and DGNN. The experimental results demonstrate that our optimized sliding-window-based training brings 0.46\% to 6.31\% accuracy improvements. Compared with their open-sourced implementations on PyTorch, \oursys achieves speedup ranging from 1.48X to 5.87X.  

\section{PRELIMINARIES}


\subsection{Graph Neural Networks}

GNNs operating on graph-structured data aim to capture the topology and feature information simultaneously. Given the input graph $G=(V,E)$ along with the features of all nodes $z^{(0)}$, GNNs stack multiple graph propagation layers to learn a representation for each node. In each layer, GNNs aggregate the neighbor information from the previous layer, following an AGGREGATE-UPDATE computation pattern:
\begin{align}
    h_{v}^{(l)} &=\texttt{AGGREGATE}\big(\{z_u^{(l-1)}\mid (u,v)\in E\},W_a^{(l)}\big),\\
    z_v^{(l)}&= \texttt{UPDATE}\big(z_v^{(l-1)}, h_{v}^{(l)}, W_u^{(l)}\big),
\end{align}
where $z_v^{(l)}$ represents the embedding of node $v$ in the $l$-th layer. The \texttt{AGGREGATE} operation collects and aggregates the embedding of $v$'s incoming neighbors to generate the intermediate aggregation result $h_{v}^{(l)}$. The \texttt{UPDATE} operation uses $h_{v}^{(l)}$ and the $(l-1)$-th layer embedding of $v$ to compute node $v$'s new embedding in the $l$-th layer, where $W_a^{(l)}$ and $W_u^{(l)}$ are model parameters in the $l$-th layer. 
\subsection{Graph Streams}



A graph stream is used to abstract the dynamic behavior of a graph over time. It can be formalized as a pair $(G^{(0)}, Evt)$, where $G^{(0)}=(V^{(0)}, E^{(0)})$ is the initial state of a graph at time $t_0$ and $Evt$ is a streaming of graph update events $\{e[0],e[1],e[2],\ldots \}$ ordered by occurrence time. Each update event can be represented as a tuple $(u, v, t, T)$ where $u$ and $v$ are the nodes involved in the event (an edge update involves two nodes and a node update only involves a single node), $t$ is the event occurred timestamp, and $T$ indicates the type of update operations, such as node/edge's addition/deletion and the update of node/edge's feature. 

Another commonly used method for modeling a dynamic graph is to represent it as a sequence of static graph snapshots, denoted as $G = (G^{(t_0)}, G^{(t_1)}, G^{(t_2)},\cdots)$. Each snapshot $G^{(t_i)}=(V^{(t_i)}, E^{(t_i)})$ denote the graph consisting of nodes $V^{(t_i)}$ and edges $E^{(t_i)}$ formed before time ${t_i}$. Each snapshot $G^{(t_i)}$ is created by incorporating dynamic information $\Delta G^{(t_i)}$ between time intervals ${t_{(i-1)}}$ and ${t_{(i-1)} + \Delta t}$ on the previous snapshot $G^{(t_{i-1})}$.

This snapshot-based method cannot represent the complete evolution process of a graph \cite{Hawkes_KDD_2018, RLDGSurvey_2020, TGN_2020_arXiv, Sparse-Dyn}. A snapshot only represents the structure at one particular time period, which can result in the loss of dynamic information. For example, if there are two update operations to add and then delete an edge between two nodes, the snapshot can only record that there is no edge between them, without capturing the dynamic change. In addition, this method cannot capture temporal dependency information within time intervals $\Delta t$  because the temporal dependency of intra-snapshot updates is not maintained. The temporal dependency information is important for modeling and prediction of dynamic graphs \cite{CTDG_WWW_2018}. For example, 
in user-item networks, recent purchases and browsing records are more likely to represent a user's latest favor, which can provide more accurate real-time recommendations for this user \cite{JODIE_arxiv_2018}. Compared with the snapshot-based method, graph streams can record all dynamic information of a graph and the temporal dependency between them, so graph streams are more general.  
Therefore, in this work, we focus on dynamic GNNs on graph streams.

\vspace{-0.05in}
\subsection{Dynamic GNN Training Abstraction}

\let\oldnl\nl
\newcommand{\nonl}{\renewcommand{\nl}{\let\nl\oldnl}}

\begin{algorithm}[ht]
    \footnotesize
    \caption{Event-based Dynamic GNN Training (Per Epoch) 
    }
    \label{alg:dynamic_train}  
    \SetKwFunction{UpdateGraph}{UpdateGraph}
    \SetKwFunction{UpdateEmb}{UpdateEmb}
    \SetKwFunction{NbrAgg}{NbrAgg}
    \SetKwFunction{Update}{Update}
     \SetKwFunction{PropUpdate}{PropUpdate}
    \SetKwFunction{Propagate}{Propagate}
    \SetKwFunction{PeerUpdate}{PeerUpdate}
    \SetKwFunction{Predict}{Predict}
    \SetKwFunction{GetSubgraph}{GetSubgraph}
    \SetKwFunction{Aggregate}{Aggregate}
    \SetKwInOut{Input}{Input} \SetKwInOut{Output}{Output}
    \Input {Initial parameters $W=\{W_{a}, W_{u}, W_{p}\}$ for aggregation, update, and propagation, respectively, initial graph $G^{(0)}=(V^{(0)}, E^{(0)})$, initial node embeddings $Z^{(0)}=\{z_v^{(0)} | \ v \in V^{(0)}\}$, a sequence of events $\{e[1], e[2],\ldots,e[n]\}$ where $e[i]=(u,v,t^i,T)$, and a set of training sample labels $Y$}
    \Output{Updated parameters $W'=\{W'_{a}, W'_{u}, W'_{p}\}$, updated graph $G^{(n)}$, and updated node embeddings $Z^{(n)}$}
        \nonl \underline{ \textbf{Forward:}}\\
        \For{ $e[i] \leftarrow e[1]$ \KwTo  $e[n]$}
        {
        $\{u,v,t^i,T\}=parse(e[i])$\;
    	$G^{(i)}$ = \UpdateGraph\big($G^{(i-1)}$, $(u,v)$, $T$\big)\;
        $\mathcal{E}_{e}(V_{e},E_{e})$ = \GetSubgraph\big($G^{(i)}$, $(u,v)$\big)\;
        $\Delta t_{u}$ = $t^{i}-t_{u}$;  $\Delta t_{v}$ = $t^{i}-t_{v}$; $t_u=t^i; t_v=t^i$\; 
        $h_{u}$ = \Aggregate \big($\{z_w^{(i-1)}\mid (w,u)\in E_{e}\}, W_{a}\big)$\;
        $h_{v}$ = \Aggregate \big($\{z_w^{(i-1)}\mid (w,v)\in E_{e}\}, W_{a}\big)$\;
         $z_{u}^{(i)}$ = 
        \UpdateEmb\big($h_{u}$, $z_{u}^{(i-1)}$, $h_{v}$,   $\Delta t_{u}$, $T$, $W_{u}$\big)\;
         $z_{v}^{(i)}$ = 
        \UpdateEmb\big( $h_{v}$, $z_{v}^{(i-1)}$, $h_{u}$, $\Delta t_{v}$, $T$, $W_{u}$\big)\;
       
       \For{ each $w \in V_{e}\setminus\{u,v\}$} {
            $z_w^{(i)}=$\PropUpdate($z_w^{(i-1)}$, $z_u^{(i)}$, $z_v^{(i)}$, $W_p$)
        
        }
        \For{ each $w \in V^{(i)}\setminus V_{e}$} {$z_w^{(i)}=z_w^{(i-1)}$
        
        }
        $\hat{y}^{(i)}$ = \Predict$\big(z_{u}^{(i)},z_{v}^{(i)}\big)$\ \eat{\wqg{How do you generate $n$ $\hat{y}$}};
    }
    $\mathcal{L} = loss\_func\big(\{\hat{y}^{(1)},\hat{y}^{(2)},\ldots,\hat{y}^{(n)}\}, Y\}\big)$;
        \newline \underline{\textbf{Backward:}}\\
        $\{\nabla W_{a}, \nabla W_u, \nabla W_p\} \leftarrow \mathcal{L}.backward()$; \\
        update $\{W'_{a}, W'_{u}, W'_{p}\}$ based on $\{\nabla W_{a}, \nabla W_u, \nabla W_p\}$;
\end{algorithm}




We find that the dynamic GNN models \cite{Hawkes_KDD_2018, JODIE_arxiv_2018, DyRep_ICLR_2019, MMDNE_CIKM_2019,  TGN_2020_arXiv, DyGNN_SIGIR_2020, TGAT_2020_arXiv, APAN_SIGMOD_2021, FDGNN_arxiv_2022} designed for graph streams adopt the traditional batched training mode to learn parameters. Each batch consists of a sequence of historical events $\{e[1], e[2],\ldots,e[n]\}$. The learning process can be abstracted as an event-based dynamic GNN training process, which is summarized in Algorithm \ref{alg:dynamic_train}. 

It starts from a initial state with graph $G^{(0)}$ and the corresponding node embedding $Z^{(0)}$ at time $t_0$. 
During training, the events are processed sequentially chronologically (Line 1-14).
First, the algorithm updates the graph topology based on the event nodes and event type of the currently scheduled event (Line 2-3) and gets the event-affected subgraph (Line 4). The computation of an event involves a subgraph, including the event nodes and their neighbors. Neighbors typically consist of one-hop neighbors, while a few algorithms consider neighbors up to $k$-hop neighbors such as JODIE \cite{JODIE_arxiv_2018}, TGN \cite{TGN_2020_arXiv}. Next, the algorithm computes the time intervals 
of the event nodes $u$, $v$ and updates their time points, respectively (Line 5). $\Delta t_{u}$ and $\Delta t_{v}$ represent the time elapsed since $u$’s previous interaction and $v$’s previous interaction, respectively, and are used as input to indicate the frequency of their interaction. Then, the algorithm computes the aggregated neighborhood embedding for each event node (Line 6-7). The aggregated neighborhood embeddings capture rich structural information, which is key to any representation learning task over graphs. 
The algorithm then computes the new embeddings $\{z_{u}^{(i)},z_{v}^{(i)}\}$ for the event nodes based on the aggregated neighbor representation $\{h_u, h_v\}$, old embeddings $\{z_{u}^{(i-1)}, z_{v}^{(i-1)}\}$ and the time interval $\{\Delta t_{u}, \Delta t_{v}\}$  (Line 8-9). 

The updated embeddings are then propagated to their neighbors (Line 10-11). This step is specific to dynamic GNN diffusion models like  DGNN \cite{DyGNN_SIGIR_2020}.
The updated event node embeddings are used as the input for downstream tasks such as link prediction and node classification (Line 14). The set of predicted labels $\hat{y}^{(i)}$ and the set of ground-truth labels $Y$ are used to calculate the loss value after executing $n$ events (Line 15). The link prediction task refers to predicting new edges between nodes based on historical events, typically without ground-truth labels. A common approach is to use negative sampling to construct negative samples and take the edge between event nodes as the positive sample. After generating positive and negative samples, the training of link prediction can be transformed into a binary classification problem.
The parameters of the model are updated based on the loss (Line 16-17).
Compared with traditional GNNs, learning dynamic node embeddings is more challenging 
due to complex time-varying graph structures.
\begin{algorithm}[t]\footnotesize
    \caption{Incremental Training with Sliding Window}
    \label{alg:incremental_learning}  
    \SetKwFunction{GetCurrentGraphInfo}{GetCurrentGraphInfo}
    \SetKwFunction{DyGNN}{DyGNN}
    \SetKwFunction{SplitData}{SplitData}
    \SetKwInOut{Input}{Input} \SetKwInOut{Output}{Output}
    \Input {$W^{(0)}$, $G^{(0)}$, $Z^{(0)}$, $\{e[1], e[2],\ldots,e[m]\}$, $Y$, fixed window size $s$, and stride $d$}
    \Output{$W^{(m)}$, $G^{(m)}$, and $Z^{(m)}$} 
    $i \leftarrow 0$;  \enspace \textcolor{gray}{//window start position}\\
    \While{$i\leq m$}
    {
        $j=(i+s<m)?(i+s):m$; \enspace \textcolor{gray}{//window end position}\\
        $\{W^{(j)}, G^{(j)}, Z^{(j)}\}=\DyGNN\big(W^{(i)}, G^{(i)}, Z^{(i)}$, $\{e[i],e[i+1]$, $\ldots, e[j]\}, Y\big)$; \enspace \textcolor{gray}{//Algorithm 1}\\
        $i \leftarrow i + d$; \enspace \textcolor{gray}{//window start position in next round}\\
    }  
\end{algorithm}

\subsection{Limitations of Existing Frameworks}
In many applications, a graph stream is a sequence of events that grows over time. To keep the model fresh, continuous training is crucial. However, existing dynamic GNN models are mainly implemented on general DNN or GNN frameworks \cite{Tensorflow_OSDI_2016, Pytorch_NIPS_2019, DGL_arXiv_2019}. They only support offline training based on events within a past time window because they lack an effective data structure to support the evolving graph topology. TGL \cite{TGL_VLDB_2022} is a training framework designed for temporal GNNs, but it is also an offline training framework that requires storing the entire dynamic graph statically before training. Therefore, TGL needs to obtain all event information in advance to build the graph structure, which loses the flexibility to support dynamic graph updates.
Additionally, TGL adopts an intra-batch parallel method, which breaks temporal dependencies between events and reduces model accuracy. To address these limitations, we propose \oursys, the first dynamic GNN framework supporting graph streaming training. We also propose a parallel execution scheme that guarantees temporal dependencies between events to accelerate model training.

\section{Dynamic GNN Training with Sliding Window}
\label{sec:learning_mode}

In this section, we propose an event-based training mode with a sliding window, which helps capture the spatial and temporal dependencies between events in a timely manner. Furthermore, we improve the basic sliding window method by adaptively adjusting the window size, aiming to include all relevant events in a window.

\subsection{Incremental Mode with Sliding Window}
\label{sec:incrementalmode}
In a streaming scenario, we assume update events are coming continuously. The key design to support training in this scenario is that we design a sliding window to select a set of training events chronologically from the input event stream. This training method is a continuous process until all events are trained or can be designed as an online process continuously accepting new update events. Algorithm \ref{alg:incremental_learning} depicts the training process on a finite event stream $\{e[1], e[2],\ldots,e[m]\}$ given the initial model parameters $W^{(0)}$, the initial graph structure $G^{(0)}$, and the initial node embeddings $Z^{(0)}$.

There are two key parameters to control the sliding window, including the 
window size $s$ and the stride $d$. For example, in Figure \ref{fig:spatial_locality} (a), the 
size $s=4$ and the stride $d=2$. The stride $d$ determines how far the window slides each time. The setting of $d$ needs to balance training efficiency and retaining information. Specifically, when $d$ is larger, the model trains faster but loses more information. The window size $s$ determines how much data can be fed into the model. The naive idea is to set the window size to a fixed value, and we have a fix-sized event set from the event stream each time. 

The algorithm first determines the start and end positions ($i$ and $j$)  of a window. Then the sequence of events in this window as input, \ie $\{e[i],\ldots, e[j]\}$, are fed into the dynamic GNN model \texttt{DyGNN} as described in Algorithm \ref{alg:dynamic_train}. The model parameters $W^{(i)}$, the graph topology $G^{(i)}$, and the node embeddings $Z^{(i)}$ at time $t_i$ are also the input of \texttt{DyGNN}, which will output the updated model parameters $W^{(j)}$, the updated graph topology $G^{(j)}$, and the updated node embeddings $Z^{(j)}$. Then the window slides with a stride $d$ to obtain the start position and end position of the next window. We repeat the above process until meeting the stream end position $m$. For an infinite stream, it will keep receiving new events and launch the \texttt{DyGNN} for a new window of events continuously.

\subsection{Adaptive Adjustment of Window Size}
\label{sec:slidingwindow}
\begin{figure}[t]
\vspace{-0.18in}
  \centering
  \includegraphics[width=\linewidth]{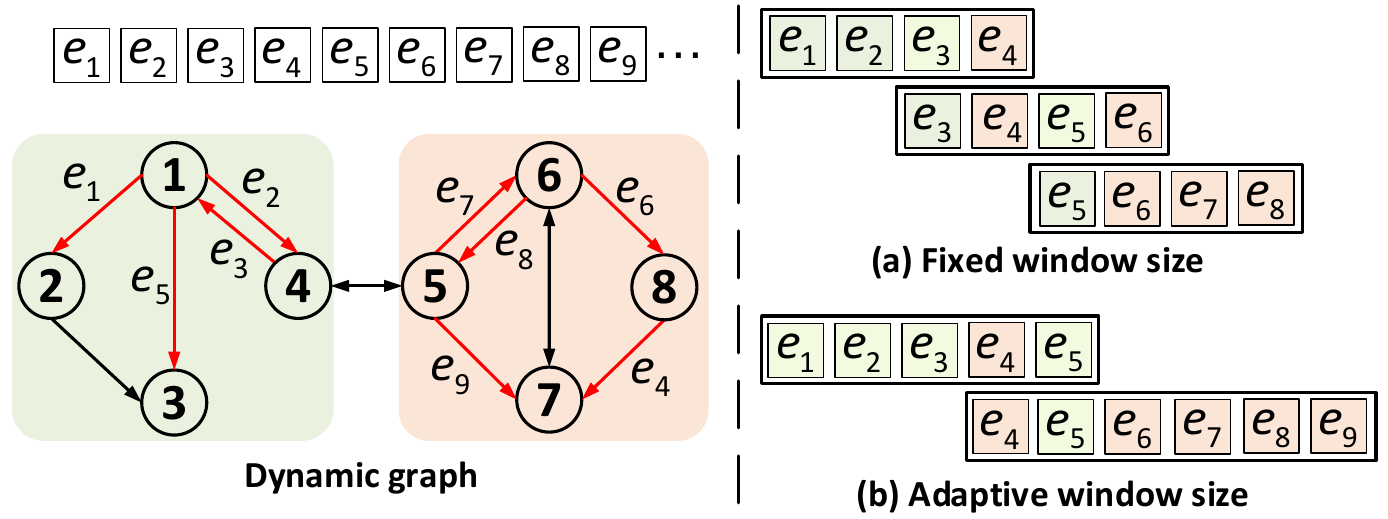}
  \vspace{-0.30in}
  \captionsetup{font=small}
  \caption{Fixed window vs. adaptive window.}
  \label{fig:spatial_locality}
  \vspace{-0.16in}
\end{figure}

The setting of the window size $s$ is a key issue. If $s$ is too small, the window contains little information. If $s$ is too large, the window may also introduce noise information, disturbing inference performance \cite{UBR4CTR_SIGIR_2020}. 
We notice that there is spatial locality among continuous events. A continuous segment of events is usually concentrated in a local area in the graph. For example, in a user-item network, a person may focus on browsing and purchasing certain items in a certain period of time, which will generate continuous interaction events. The model needs to encode the information of these highly related events into the user's embedding in time, so this sequence of correlated events should all be contained in a window, which cannot be satisfied by a fixed window size. 
As shown in Figure \ref{fig:spatial_locality}, there are 9 events where event set $\{e_1 , e_2, e_3, e_5\}$ occurs in the green area and event set $\{e_4 , e_6, e_7, e_8, e_9\}$ occurs in the orange area. However, if the window size is a fixed value of 4, the model cannot learn the spatial locality relations among these events. Figure \ref{fig:spatial_locality} (a) shows that none of the three windows can contain one of the event sets. Therefore, the window size should be dynamically adjustable to capture the spatial locality, as shown in Figure \ref{fig:spatial_locality} (b).
Therefore, we propose an algorithm to adaptively adjust the window size for capturing the spatial locality as shown in Algorithm \ref{alg:auto_window}.





\begin{algorithm}[t]\footnotesize
    \caption{Determination of a Window of Events}
    \label{alg:auto_window}  
    \SetKwFunction{GetEvent}{GetEvent}
    \SetKwInOut{Input}{Input} \SetKwInOut{Output}{Output}
    \Input {Event stream $\{e[i], e[i+1], \ldots, e[m]\}$, minimum window size $L$, and maximum window size $H$}
    \Output{A window of events $\{e[i],e[i+1],\ldots,e[i+s]\}$} 
    $s\leftarrow 0$; \enspace \textcolor{gray}{//window size}\\ 
    $nodes\leftarrow\emptyset$;   \enspace \textcolor{gray}{//a node set for capturing event dependencies}\\
    \While{$s \leq H$}
    {
        $j\leftarrow i+s$; \enspace \textcolor{gray}{//window end position}\\
        $\{u,v,t^j,T\}=parse(e[j])$\;
        \If{$s \leq L$}{
            $nodes\leftarrow nodes \cup \{u,v\}$; \enspace \textcolor{gray}{//not enough events}\
        }
        \ElseIf{$\{u, v\} \cap nodes \neq\emptyset $}{
            $nodes\leftarrow nodes \cup \{u,v\}$; \enspace \textcolor{gray}{//spatial dependent event}\
        }
        \Else{
            \textbf{break};\\        
        }
        $s\leftarrow s+1$; \enspace \textcolor{gray}{//updated window size}\
    }  
\Return $\{e[i],e[i+1],\ldots,e[i+s]\}$
\end{algorithm}

The idea of Algorithm \ref{alg:auto_window} is to iteratively judge whether there is a correlation between the following event outside the window and the events already contained in the window. 
The inputs include an event stream $\{e[i], e[i+1], \ldots, e[m]\}$, minimum window size $L$, and maximum window size $H$. $L$ and $H$ are two hyperparameters that limit the number of events the window can contain. We adopt these two hyperparameters to control the amount of events input into the model training each time. If there is no association between a sequence of events, we use the minimum size $L$ to ensure the least amount of training data. If there are many associations among sequential events, this results in more and more events being added to the window, and the window can be infinitely long. Therefore, we use the maximum size $H$ to limit the maximum length of the window. The output 
is a window of events $\{e[i],e[i+1],\ldots,e[i+s]\}$.

The algorithm first initializes the set $nodes$ and $s$, where $nodes$ represents a set of event nodes already in the window and $s$ represents the number of events already added (Line 1-2). Then the algorithm  sequentially detects whether each event has spatial locality to events already in the window  (Line 3-12).  We get an event $e[j]$ from the event stream based on the index $j$ each time (Line 4-5). Next, we perform a judgment for the event $e[j]$. If the number of added events in the window is less than the minimum size $L$, this event is directly added to the window, and the event nodes $\{u,v\}$ are added to the set $nodes$ (Line 6-7). Otherwise, the algorithm judges whether there is an intersection between event nodes $\{u,v\}$ of event $e[j]$ and the set $nodes$. If there is an intersection, it means a locality exists between the $e[j]$ and the events already in the window. Therefore, this event should be added to the window (Line 8-9). If neither condition is met, break out of the loop (Line 10-11). If the number of events in the window reaches the maximum size $H$, break out of the loop (Line 3). Finally, we can get the range $[i, i+s]$ of events in this window, and then we return the window (Line 13).

Based on Algorithm \ref{alg:auto_window}, we can put a segment of events with spatial locality into a window so that the model can learn the locality between events. This algorithm can replace the fixed window size 
in Algorithm \ref{alg:incremental_learning}.
{\color{black}{Another method is to check whether there is an intersection between consecutive events based on the event subgraph. However, this method makes dependency analysis more time-consuming. Moreover, most dynamic GNNs only update event nodes and do not update their neighbors. Therefore, it is reasonable to use Algorithm \ref{alg:auto_window} to determine the window.}}
\section{Dependency-Graph-Driven Event Parallelizing}
\label{sec:parallel}



\begin{figure*}[t]
\vspace{-0.5in}
  \centering  \includegraphics[width=\linewidth]{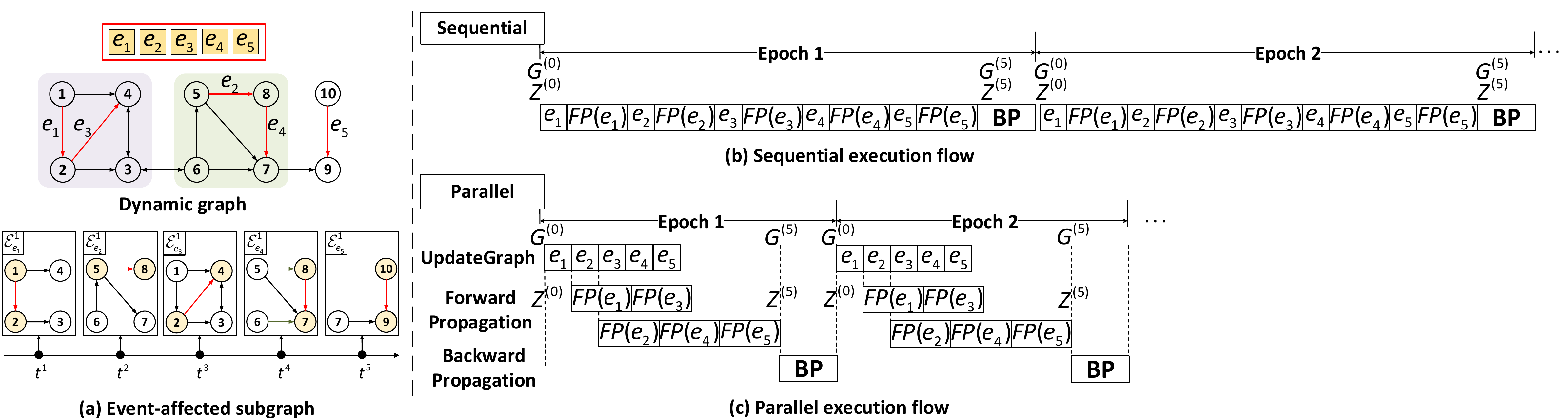}
  \vspace{-0.22in}
  \captionsetup{font=small}
  \caption{Execution flow comparison of sequential execution events and parallel execution events. The red arrows represent the newly added edge. FP represents forward propagation, while BP represents backward propagation. Figure (a) is the 1-hop event-affected subgraph $\mathcal{E}^{1}_e$ of event $e_1$, $e_2$, $e_3$, $e_4$, $e_5$. The purple area is the affected subgraph of events $e_1$ and $e_3$, and the green area is the affected subgraph of events $e_2$ and $e_4$.}
  \label{fig:exeflow}
  \vspace{-0.10in}
\end{figure*}

When training the dynamic GNN model with a window, the events in the window are executed chronologically, as shown in Algorithm \ref{alg:dynamic_train}. For example, in Figure \ref{fig:exeflow} (a), a window contains five events.  
Figure \ref{fig:exeflow} (b) illustrates that the model executes these events sequentially, leading to a long training time. However, we notice that the computation of an event only involves a subgraph, typically consisting of the event nodes and their neighbors. If there is no intersection of subgraphs between two events, they can be executed in parallel.  
For example, in Figure \ref{fig:exeflow} (a), the purple area is the subgraph of events $e_1$ and $e_3$, and the green area is the subgraph of events $e_2$ and $e_4$. The two areas are disjoint, so events $e_1$ and $e_3$ can be executed in parallel with events $e_2$ and $e_4$. 
This fact brings the opportunity of parallelism in event processing. In this section, we first analyze the conditions of event parallelism and give related definitions. We then propose an algorithm to automatically analyze the dependencies of events and identify the events that can be executed in parallel.

\vspace{-0.10in}
\subsection{Dependency Analysis of Event Processing}


The parallel processing of events can be explained by answering two questions.

\textbf{Question 1: Why are there dependencies among events?}

If there are no read-write conflicts between two operations, the two operations can be processed in parallel. Therefore, we first define an event’s read set and write set. As shown in Algorithm \ref{alg:dynamic_train}, An event $e=(u, v, t, T)$ involves event nodes $\{u, v\}$ and their $k$-hop neighbors where $k$ is determined by the model.
The computation of an event requires reading and writing information about these nodes from the graph topology and node embeddings. For example, $\Aggregate$ requires reading the node embeddings of neighbors, while $\UpdateEmb$ requires updating the embeddings of event nodes.  
As shown in Figure \ref{fig:exeflow} (a), the event $e_1$ involves event nodes 1, 2 and their one-hop neighbors 3, 4. Therefore, $\Aggregate$ requires reading the embeddings of neighbors 3, 4, while $\UpdateEmb$ requires writing the embeddings of event nodes 1, 2.





Based on the above analysis, we define \textbf{Event-Affected Subgraph} and \textbf{Event-Triggered Update Set} to denote the read range and the write range of an event, respectively. 
\begin{definition}
\label{def:EASG}
\textbf{Event-Affected Subgraph.}
For an event $e=(u, v, t, T)$, it generally needs to read the information of the subgraph composed of the $k$-hop neighbors of the event nodes $\{u,v\}$. We refer to this subgraph as the $k$-hop Event-Affected Subgraph $\mathcal{E}^{k}_e (V^k_{e},E^k_{e})$, where $V^k_e = \{\mathcal{N}_u^{k} \cup \mathcal{N}_v^{k}\}$ and $E^k_e$ is the set of edges between them.
\end{definition}

\begin{definition}
\label{def:update}
\textbf{Event-Triggered Update Set.}
After processing an event $e=(u, v, t, T)$, the embeddings of the nodes around $u$ and $v$ need to be updated. This set of nodes is referred to as the event-triggered update set and is denoted by $\mathcal{U}_e$.
\end{definition}



For example in Figure \ref{fig:exeflow} (a), the 1-hop event-affected subgraph of event $e_1$ is $\mathcal{E}^{1}_{e_1}$ and $V^1_{e_1} = \{1, 2, 3, 4\}$. Its event-triggered update set is $\mathcal{U}_{e_1}=\{1,2\}$. 
The 1-hop event-affected subgraph of event $e_2$ is $\mathcal{E}^{1}_{e_2}$ and $V^1_{e_2} = \{5,6,7, 8\}$. Its event-triggered update set is $\mathcal{U}_{e_2}=\{5,8\}$. 
Based on  Definition \ref{def:EASG} and Definition \ref{def:update}, we define a dependency between two events as having a read-write conflict as follows.


\begin{definition}
\label{def:dep}
\textbf{Dependencies between events. }
For any two events $e_i=(u_i, v_i, t_i, T_i)$ and $e_j=(u_j, v_j, t_j, T_j)$ where $t_i < t_j$, we have their corresponding event-affected subgraphs, i.e., $\mathcal{E}_{e_i}^k(V_{e_i}^k, E_{e_i}^k)$ and $\mathcal{E}_{e_j}^k(V_{e_j}^k, E_{e_j}^k)$, and event-triggered update sets, i.e., $\mathcal{U}_{e_i}$ and $\mathcal{U}_{e_j}$. If $V_{e_j}^k\cap \mathcal{U}_{e_i}\ne \varnothing$, we say that 
event $e_j$ depends on event $e_i$, i.e., $e_j\Rightarrow e_i$.
\end{definition}

For example, in Figure \ref{fig:exeflow} (a), the 1-hop event-affected node set of event $e_3$ is $V^1_{e_3} = \{1, 2, 3, 4\}$. The event-triggered update set of $e_1$ is $\mathcal{U}_{e_1}=\{1,2\}$. $V^1_{e_3} \cap \mathcal{U}_{e_1} = \{1, 2\}$, which is not an empty set, so event $e_3$ depends on event $e_1$, marked as $e_3 \Rightarrow e_1$. 

The dependency relationship $e_j\Rightarrow e_i$ indicates that event $e_j$ must wait for event $e_i$ to complete before executing. Similarly, if another event $e_k$ depends on event $e_j$, event $e_k$ must also wait for event $e_j$ to complete. Therefore, Event $e_k$ must also wait for event $e_i$. Event $e_k$ further depends on event $e_i$, resulting in a dependency chain, ie. $e_k \Rightarrow e_j \Rightarrow e_i$. The dependencies between events satisfy the transitivity theorem. The theorem is as follows.


\begin{theorem}
\label{def:pass}
\textbf{Dependency transitivity. } 
For any three events $e_i$, $e_j$ and $e_k$, if $e_j \Rightarrow e_i$, $e_k \Rightarrow e_j$, then $e_k \Rightarrow e_i$.
\end{theorem}


For a dependency chain $e_k \Rightarrow e_j \Rightarrow e_i$, non-adjacent events do not necessarily satisfy Definition \ref{def:dep}. For example, in Figure \ref{fig:exeflow} (a), the event-triggered update set of $e_2$ is $\mathcal{U}_{e_2}=\{5,8\}$. The event-affected node set of event $e_4$ is $V^1_{e_4} = \{5, 6, 7, 8\}$. $\mathcal{U}_{e_2} \cap V^1_{e_4}  = \{5, 8\}$, which is not an empty set, so event $e_4$ depends on event $e_2$. 
Similarly, the event-triggered update set of $e_4$ is $\mathcal{U}_{e_4}=\{7,8\}$. The event-affected node set of  event $e_5$ is $V^1_{e_5} = \{7, 9, 10\}$. $\mathcal{U}_{e_4} \cap V^1_{e_5}  = \{7\}$, which is not an empty set, so event $e_5$ depends on event $e_4$. Therefore, the dependencies of these three events form a dependency chain, ie. $e_5 \Rightarrow e_4 \Rightarrow e_2$. However, event $e_2$ has no read-write conflict with event $e_5$, ie. $\mathcal{U}_{e_2} \cap V^1_{e_5}  = \varnothing$. Therefore, Definition \ref{def:dep} is a sufficient condition for the dependency. Two adjacent events in a dependency chain always satisfy Definition \ref{def:dep}, so we refer to this dependency as a direct dependency.

\textbf{Question 2: Under what conditions can events be parallelized?}



Definition \ref{def:dep} and Theorem \ref{def:pass} indicate that if two events are not in the same dependency chain,  they can be executed in parallel.
As shown in Figure \ref{fig:exeflow}, the 1-hop event-triggered update set of event $e_1$ is $\mathcal{U}_{e_1}=\{1,2\}$. The event-affected node set of event $e_3$ is $V^1_{e_3} = \{1,2,3, 4\}$. Since $\mathcal{U}_{e_1} \cap V^1_{e_3} = \{1, 2\}$, which is not an empty set, so event $e_3$ depends on event $e_1$. Therefore, these two events form a dependency chain, ie. $e_3 \Rightarrow e_1$. Similarly,  event $e_2$, event $e_4$ and event $e_5$  form another dependency chain, ie. $e_5 \Rightarrow e_4 \Rightarrow e_2$. Therefore, the events of these two chains can be executed in parallel, as shown in Figure \ref{fig:exeflow} (c).
Parallel training can obviously speed up model training.
However, existing frameworks \cite{Pytorch_NIPS_2019, Tensorflow_OSDI_2016, DGL_arXiv_2019} cannot analyze the dependencies between events and can only process these events sequentially, causing degraded performance. 
Therefore, we propose an algorithm to automatically analyze event dependencies, achieving parallel execution between events while strictly guaranteeing temporal dependency.

\vspace{-0.05in}
\begin{figure}[ht]
\vspace{-0.06in}
  \centering
  \includegraphics[width=\linewidth]{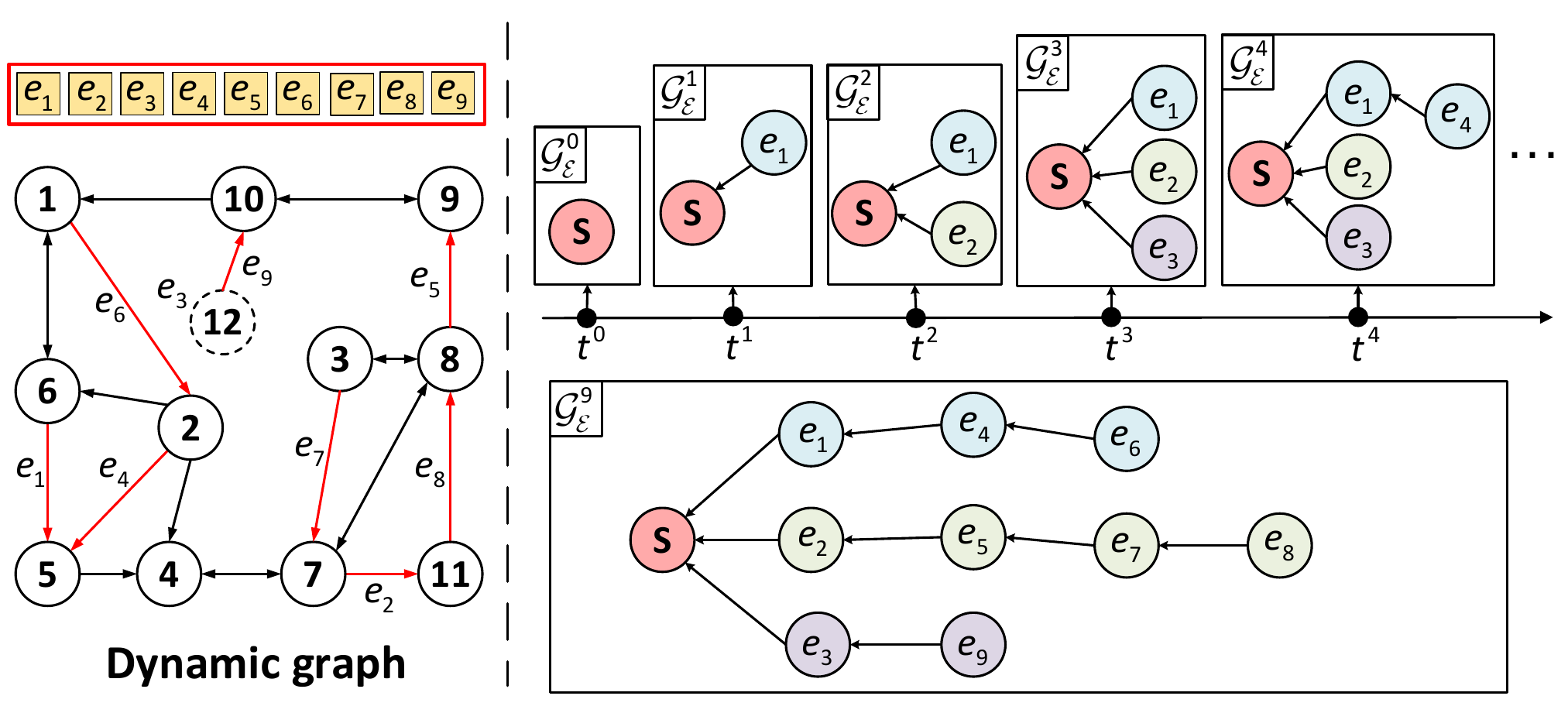}
  \vspace{-0.24in}
  \captionsetup{font=small}
  \caption{Event dependency graph. The red arrows represent the newly added edge, and the dotted line represents the newly added node. The upper right corner is the evolved Event Dependency Graph. The lower right corner is the complete event dependency graph, where nodes of the same color represent a dependency chain.}
  \label{fig:EASG}
  \vspace{-0.05in}
\end{figure}

\subsection{Parallelization based on Event Dependency Graph}



We define the \textbf{Event Dependency Graph} to represent dependencies between events as follows.


\begin{definition}
\label{def:dep_graph}
\textbf{Event Dependency Graph. }
An event dependency graph is denoted as $\mathcal{G}_{\mathcal{E}} = (\mathcal{V}, \mathcal{E})$. Here $\mathcal{V}$ is a set of nodes where each node represents an event. $\mathcal{E} \subseteq \mathcal{V} \times \mathcal{V}$ is a set of edges where each edge represents a dependency. For example, a directed edge $(e_j, e_i)$ indicates that event $e_j$ directly depends on event $e_i$, i.e., $e_j\Rightarrow e_i$.
\end{definition}

We introduce a super-node marked $S$ as the root node in $\mathcal{G}_{\mathcal{E}}$, and all events depend on it.
For example, in Figure \ref{fig:EASG}, a window contains nine events. Figure \ref{fig:EASG} depicts the event dependency graph $\mathcal{G}^9_{\mathcal{E}}$ of these events. For example, edge $(1, S)$ indicates that $e_1$ depends on $S$. Similarly, edge $(4, 1)$ indicates that $e_4$ depends on $e_1$, and edge $(6, 4)$ indicates that $e_6$ depends on $e_4$. Since event dependencies are transitive, $e_6$ also depends on $e_1$ and $S$. Therefore, these three events form a dependency chain, i.e., $e_6 \Rightarrow e_4 \Rightarrow e_1 \Rightarrow S$. Similarly, event set $\{e_2, e_5, e_7, e_8 \}$ forms a dependency chain, and event set $\{e_3, e_9\}$ forms another chain. Therefore, these nine events form three dependency chains.
We can execute events in parallel based on this dependency graph. 
Events can be executed in parallel if they do not belong to the same dependency chain. Therefore, the events in these three chains can be executed in parallel. To infer the event dependency graph, we propose an event dependency graph generation algorithm, as shown in Algorithm \ref{alg:evtparallel}.

\begin{algorithm}\footnotesize
\caption{Generate Event Dependency Graph}
\label{alg:evtparallel}
\SetKwFunction{IsDep}{IsDepended}
\SetKwInOut{Input}{Input}\SetKwInOut{Output}{Output}
\Input{A window of events $\{e[1], e[2],\ldots,e[s]\}$}
\Output{Event dependency graph $\mathcal{G}$}
initialize $\mathcal{G}$ with a root $e[0]$; \enspace \textcolor{gray}{//for any $e[i], e[i]\Rightarrow e[0]$}\\
\For{$e[i] \leftarrow e[1]$ \KwTo $e[s]$}
{
    \For{$e[j] \leftarrow e[i-1]$ \KwTo $e[0]$}{
        \If{$e[i]\Rightarrow e[j]$}{ 
            add event dependency $(e[i]\Rightarrow e[j])$ to $\mathcal{G}$\;
            add $e[j]$ to $e[i].deps[]$\;
        }
    }
}
\Return $\mathcal{G}$;
\end{algorithm}


In a dependency chain, the direct dependencies between two adjacent events always satisfy Definition \ref{def:dep}. Therefore, the idea of Algorithm \ref{alg:evtparallel} is to construct a dependency graph by identifying all direct dependencies. The input 
is a window of events, and the output is the event dependency graph of this window. The algorithm first initializes the dependency graph with a root event $e[0]$, where all events depend on $e[0]$ (Line 1). Next, the algorithm sequentially detects dependencies between each event $e[i]$ and the preceding events $e[i-1]$ to $e[0]$ (Line 2-6). The algorithm iteratively judge the dependencies between $e[i]$ and the events $e[i-1]$ to $e[0]$ (Line 3-6). If $e[i]$ depends on $e[j]$, we add an edge from $e[i]$ to $e[j]$ in the graph and add $e[j]$ to $e[i]$'s the dependency set $e[i].deps[]$. Finally, we can obtain the dependency graph of this window (Line 7).

Next, we can execute in parallel based on this generated dependency graph. We maintain a flag for each event indicating whether it has been completed. We use two queues, \texttt{UnExec} and  \texttt{Exec}, to manage the execution state of events. \texttt{UnExec} stores events that have not been executed, while \texttt{Exec} stores events that can be executed in parallel. First, we traverse the generated graph and put all events into \texttt{UnExec}.  Then, the scheduling thread continuously detects whether the dependency set $deps[]$ of each event in \texttt{UnExec} is empty. If the dependency set of an event is empty, it indicates that the event can be executed immediately. The scheduling thread moves the event to \texttt{Exec}. Multiple execution threads take events from \texttt{Exec} and execute them in parallel.

\section{\oursys}
\label{sec:neutronstream}

\subsection{Dynamic Graph Storage \label{sec:storage}}

\begin{figure}[t]
\vspace{-0.2in}
    \centering
    \subfloat[In-out-edge-separated indexed adjacency lists]{
	   \includegraphics[width=\linewidth]{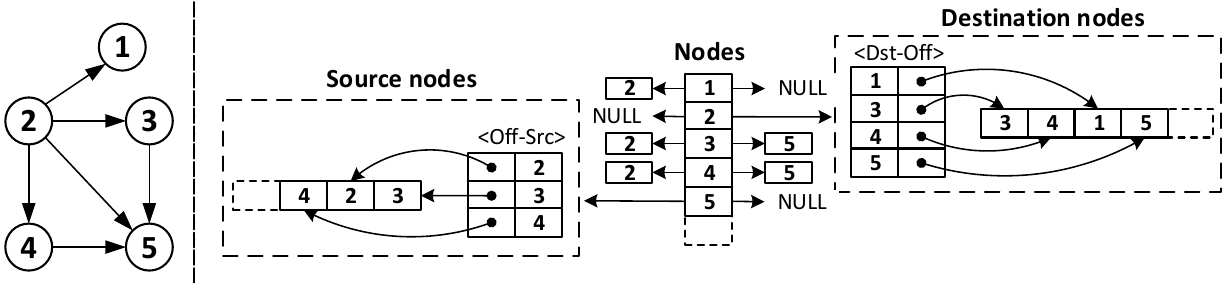}
        } 
        \vspace{-0.2in}
      \hfill
    \subfloat[Multi-version node embedding storage]{
	\includegraphics[width=0.80\linewidth]{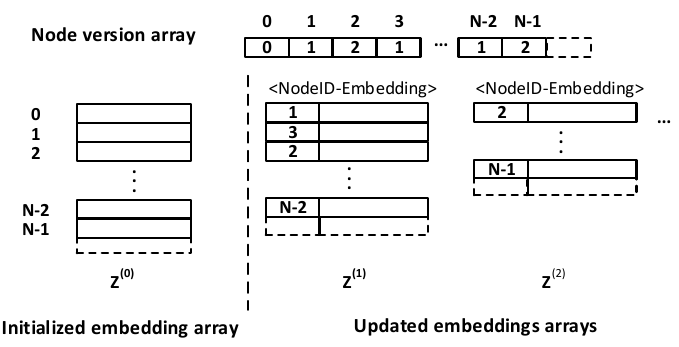}
        }
        \vspace{-0.15in}
        \captionsetup{font=small}
	\caption{Dynamic graph storage design.}
      \vspace{-0.20in}
      \label{fig:graphstrorage}
\end{figure}



As shown in Algorithm \ref{alg:dynamic_train}, the complex computations of dynamic GNNs require storage structures that can provide efficient read and write operations. In the graph streaming training process, the storage structure also needs to support dynamic updates, such as node additions and edge additions. 
However, the mature DNN/GNN frameworks \cite{Tensorflow_OSDI_2016, Pytorch_NIPS_2019, DGL_arXiv_2019} lack support for dynamic graphs, requiring users to design and manage dynamic graphs, increasing the burden. Manually implemented dynamic graphs are typically inefficient and inflexible. 
Although there are some dynamic graph storages \cite{livegraph_vldb_2020, Teseo_vldb_2021, Sortledton_sigmod_2023}, they focus on 
graph analysis workloads and do not natively support storing node vectors. Dynamic GNNs require storage structures 
supporting node embedding storage and multi-version embedding storage for backpropagation.
Therefore, we design a dynamic graph data structure \textit{in-out-edge-separated indexed adjacency lists} and a node embedding data structure \textit{multi-version node embedding} to address these requirements, as shown in Figure \ref{fig:graphstrorage}.

\Paragraph{Dynamic Graph Storage}
We propose a dynamic graph storage 
as shown in Figure \ref{fig:graphstrorage} (a). 
It efficiently and flexibly stores incoming and outgoing edges separately for each node because some models need to access both incoming and outgoing edges like DGNN \cite{DyGNN_SIGIR_2020}.

Each node has two dynamic arrays to store the incoming and outgoing edges, including source/destination node IDs, edge weights, and timestamps. Two dynamic arrays ensure that a node's incoming and outgoing edges can be visited continuously. However, arrays suffer from edge lookups by scanning for fine-grained access. For example, when an event needs to update the timestamp of a specified edge, we need to scan this array sequentially, and the time complexity 
is $O(n)$. To accelerate lookups, we maintain Key-Value pairs of $\langle NeighborID \rightarrow Offset \rangle$ for edges, indicating edge locations in arrays. 
We use Hash Table as the default index because our data structure with Hash Table provides an average $O(1)$ time complexity for each fine-grained access. Furthermore, the index does not hurt the analysis performance of traversing a node's neighbors because we can directly access the dynamic array to get all neighbors without involving the index. The structure indexes can directly locate edges without scanning, which is friendly for insertions and deletions of an edge. Since we use dynamic arrays to store nodes and edges, dynamic graph updates are naturally supported.



\Paragraph{Multi-Version Node Embedding Storage}  To support batch backpropagation, storing multiple versions of node embeddings for nodes that are updated multiple times within a batch is necessary. Figure \ref{fig:graphstrorage} (b) illustrates our proposed multi-version embedding storage structure, which optimizes storage usage and enables efficient read and write operations. The embedding storage contains three components: initialized embedding array, updated embedding arrays, and node version array.

Before training starts, the initial node embeddings are stored in a dynamic array $Z^{(0)}$, marked as version 0. The version number is incremented when the node embedding is updated. We store updated embeddings with the same version together. Each embedding is  stored as a key-value pair $\langle NodeID \rightarrow Embedding \rangle$. We use Hash Table as the default index to speed up access. The node version array records the latest node version, indicating in which dynamic array the latest embedding of each node is stored. For example, the latest version of node 0 is 0, meaning that we need to find the latest embedding in $Z^{(0)}$. Similarly, the latest version of node 2 is 2, meaning that we need to find the latest embedding in $Z^{(2)}$.


Based on this structure, we can quickly access and store the specific node embedding through two indexes with a time complexity of $O (1)$. For example, if we want to query the embedding of node 2 with version number 1. We first need to get the latest version number of node 2 according to the node ID in the node version array. We can get that the version is 2, which indicates that there is an embedding with version number 1. We can then locate the embedding in $Z^{(1)}$ according to the node ID. The multi-version node embedding storage can support efficient access and modification, avoid redundant storage and provide users with more flexibility.   

\subsection{Programming Model and APIs \label{sec:api}}

Implementing dynamic GNN models based on existing NN/GNN frameworks \cite{Pytorch_NIPS_2019, Tensorflow_OSDI_2016, DGL_arXiv_2019} requires users to implement storage structures and various complex graph operators manually, which requires high human efforts and is also tedious and error-prone. To facilitate the management of dynamic graphs, we provide a set of APIs for accessing the underlying graph storage and node embedding storage.  To enable users to focus on implementing the algorithm logic, we abstract the execution process of an event into four programming interfaces according to Algorithm \ref{alg:dynamic_train}: \texttt{\textbf{UpdateGraph}}, \texttt{\textbf{Aggregate}}, \texttt{\textbf{UpdateEmb}}, and \texttt{\textbf{PropUpdate}}. Users only need to customize these four functions to implement the execution logic of an event. The details of these APIs are listed in the following.

\begin{itemize}[leftmargin=*]
    \item \texttt{\textbf{DynGraph}} is the underlying graph storage. It provides various functions, such as adding an edge (\texttt{add\_edge(uid, vid)}), deleting an edge (\texttt{del\_edge(uid, vid)}), adding a node  (\texttt{add\_node(nid)}), getting an event subgraph (\texttt{get\_subgraph(evt)}), and querying the latest interaction time of a node (\texttt{query\_time (nid)}).

    \item \texttt{\textbf{DynEmb}} is the underlying node embedding storage. It provides various functions, such as querying a node embedding (\texttt{index (nid)}) and updating a node embedding (\texttt{update(nid, upd\_emd)}).
    \item \texttt{\textbf{UpdateGraph}} is a user-defined function that updates the graph structure based on event types, such as adding/removing edges. The parameters include the dynamic graph \texttt{dynGraph} and an event \texttt{evt}.

    \item \texttt{\textbf{Aggregate}}  is a user-defined function that computes aggregated neighbor information for an event node. The parameters include event subgraph \texttt{dynSubgraph}, the node embedding \texttt{dynEmb}, and an event node \texttt{nid}.
    
    \item \texttt{\textbf{UpdateEmb}} is a user-defined function that computes updated embedding for an event node. The parameters include event subgraph \texttt{dynSubgraph}, the node embedding \texttt{dynEmb}, an event node \texttt{nid}, aggregated neighbor information \texttt{h\_agg}, and event timestamp \texttt{time}.

     \item \texttt{\textbf{PropUpdate}} is a user-defined function that propagates an event node's embedding to neighbors. This function 
     appears in diffusion models, such as DGNN \cite{DyGNN_SIGIR_2020}. The parameters include \texttt{dynSubgraph}, 
     \texttt{dynEmb}, and an event node ID \texttt{nid}.

    
    
\end{itemize}

\begin{figure}[t]
  \centering
 \includegraphics[width=\linewidth]{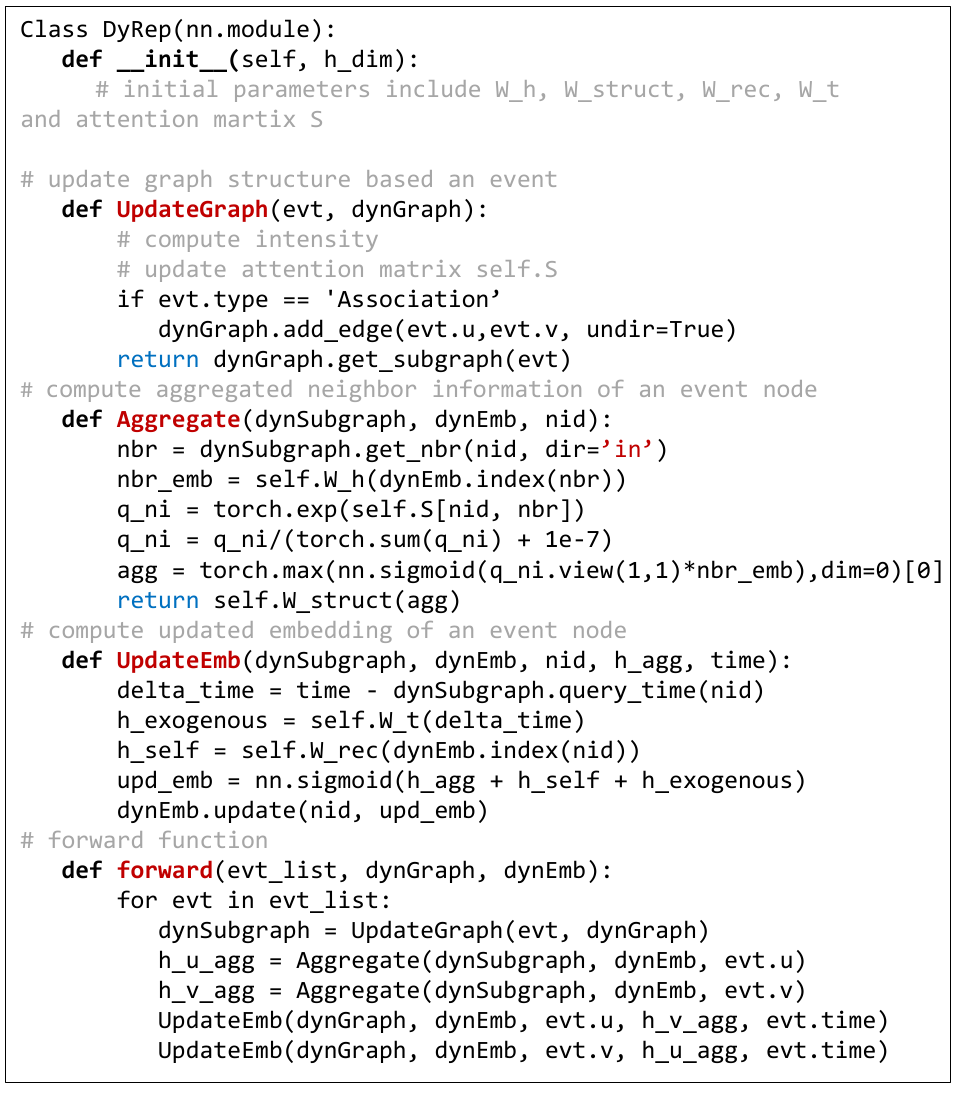}
 \vspace{-0.25in}
  \captionsetup{font=small}
  \caption{DyRep implementation with \oursys APIs.}
  \label{fig:dyrep}
  \vspace{-0.15in}
\end{figure}
We can easily represent existing dynamic GNN models with the above interfaces. For each event node, DyRep \cite{DyRep_ICLR_2019} needs to update its embedding based on three parts: Localized Embedding Propagation, Self-Propagation, and Exogenous Drive. We use \texttt{UpdateGraph} to update the graph structure based on the update event. Localized Embedding Propagation needs to aggregate neighbor information, which can be represented by \texttt{Aggregate}. Self-Propagation and Exogenous Drive refer to computations based on previous embedding information and time interval of event nodes, respectively. We can use \texttt{UpdateEmb} to compute Self-Propagation and Exogenous Drive and combine the three parts to obtain the updated event node embeddings. 
Figure \ref{fig:dyrep} shows the logic of computing event node embeddings in the Dyrep model on \oursys.
Users only need to define \texttt{\textbf{UpdateGraph}}, \texttt{\textbf{Aggregate}}, and \texttt{\textbf{UpdateEmb}}.


\vspace{-0.10in}
\subsection{Event Processing Engine and Pipelining \label{sec:eventframework}}

The Event Processing Engine is responsible for window determination and parallel training. The framework consists of four modules: Adaptive sliding window module, Event parallel analysis module, Event parallel scheduling module, and Event execution module. Each module performs different training tasks. First, the adaptive sliding window module determines a window of events from the event stream according to Algorithm \ref{alg:auto_window}. Then, the event parallel analysis module generates the event dependency graph of the window according to Algorithm \ref{alg:evtparallel}. Based on the generated dependency graph, the event parallel scheduling module generates a parallel execution plan and dispatches executable events to the event execution module. Next, the execution module utilizes the thread pool to execute events in parallel. 
The sequential nature of time-dependent events requires that events be processed sequentially, making it difficult to divide training tasks and implement pipeline optimization. Based on our event parallel approach, we can divide the training process into different tasks, making pipeline optimization possible.


\begin{figure}[t]
  \centering
  \includegraphics[width=\linewidth]{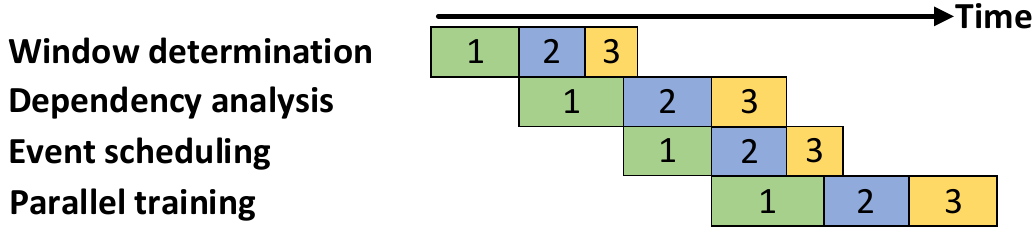}
  \vspace{-0.3in}
  \captionsetup{font=small}
  \caption{Pipeline optimization. The colors green, blue, and yellow indicate different windows of training events. 
  }
  \label{fig:pipeline}
  \vspace{-0.20in}
\end{figure}

We propose a pipelining optimization to fully use each module, as shown in Figure \ref{fig:pipeline}. The pipeline enables concurrent execution of different windows. Once the first window is obtained, the adaptive sliding window module can continue to determine the second window. After the event analysis module has generated the event dependency graph of the first window, it can continue to analyze the second window. Similarly, the event parallel scheduling module can generate a scheduling plan for the second window without waiting for the first window to complete. The design can reduce the waiting time of each module and speed up training. 
Furthermore,  dependency is important. Ignoring dependencies and executing events completely in parallel could cause the model to fail to capture dynamic information correctly and reduce accuracy. 

\begin{figure}[ht] 
\vspace{-0.15in}
\flushleft
\begin{minipage}{.62\linewidth}
    \begin{center}
    \includegraphics[width=1\linewidth]{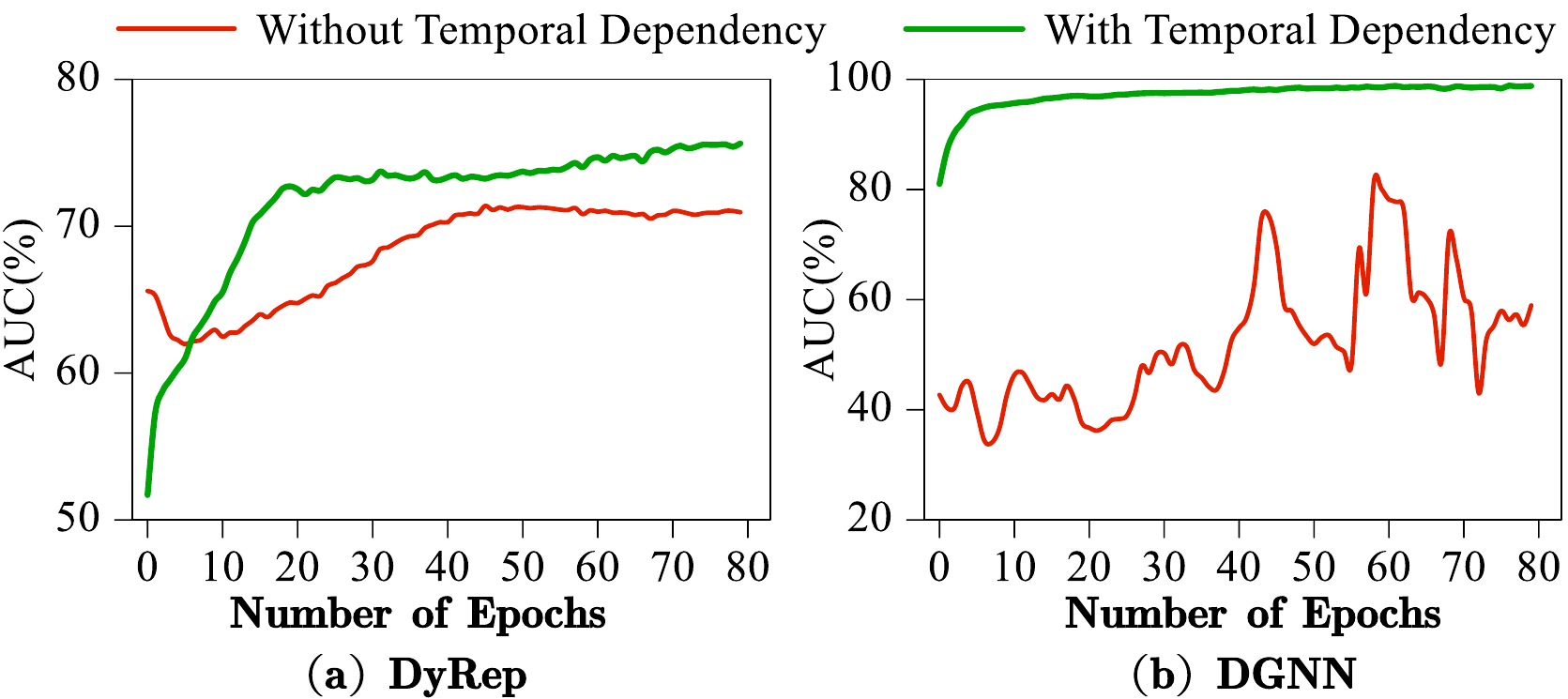}
      \vspace{-0.29in}
    \captionsetup{font=small}
    \caption{AUC Comparison with dependency and without dependency.}
    \label{fig:dep_auc}
    \end{center}
\end{minipage}
 \hspace{0.02in}
\begin{minipage}{.35\linewidth}
    \begin{center}
    \includegraphics[width=1\linewidth]{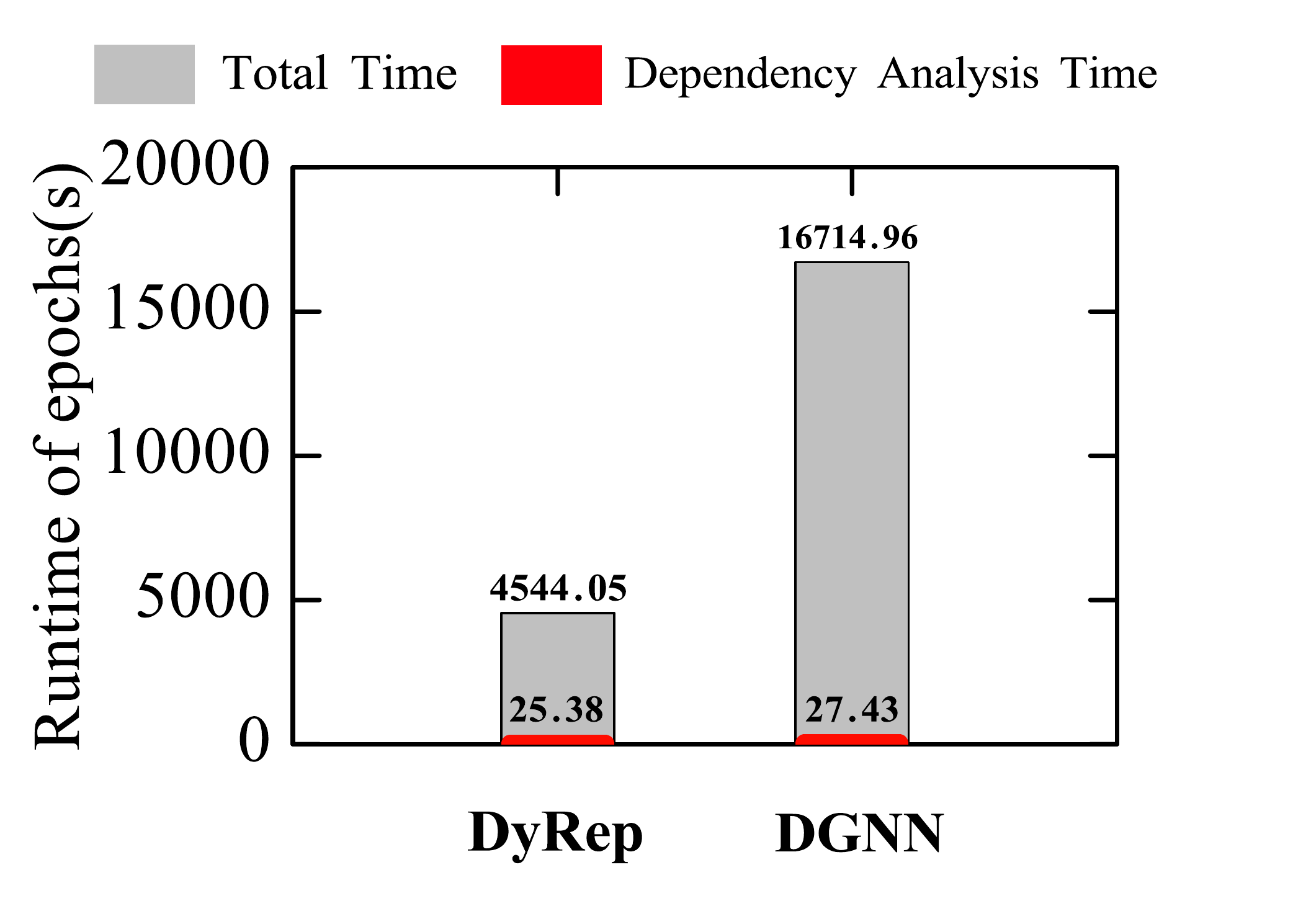}
      \vspace{-0.21in}
    \captionsetup{font=small}
    \caption{\addrevision{Ratio of dependency analysis time.}} \label{fig:dep_analysis}
    \end{center}
\end{minipage}

 \vspace{-0.2in}
\end{figure}

\Paragraph{Evaluation of the Impact on Temporal Dependency}
We evaluate the impact of temporal dependency on DyRep \cite{DyRep_ICLR_2019} and DGNN \cite{DyGNN_SIGIR_2020} with the Social dataset, as shown in Table \ref{tab:with_dep} and Figure \ref{fig:dep_auc}. With Temporal Dependency indicates that the model executes events according timestamps. We use our parallel approach to run this pattern because our 
approach can strictly guarantee dependency among events (Algorithm \ref{alg:evtparallel}). Without Temporal Dependency indicates ignoring the temporal dependency among events and executing them completely in parallel. We perform training on the first 80\% of events and testing on the final 20\%. As shown in Table \ref{tab:with_dep}, Without Temporal Dependency achieves a higher speedup. However, the speedup is limited due to conflicts in reading and writing node embeddings. As shown in Figure \ref{fig:dep_auc}, Without Temporal Dependency completely loses time information and therefore has lower accuracy. With Temporal Dependency retains accurate dependency information, enabling the model to capture dynamic information correctly and achieve higher accuracy. 
Therefore, it is necessary to maintain the dependencies among events. We also record the time of performing dependency analysis according to Algorithm \ref{alg:evtparallel}, as shown in Figure \ref{fig:dep_analysis}. The execution time of Algorithm \ref{alg:evtparallel} only takes up a small ratio of the total time, but the training can be parallelized. (Note that this time is recorded when the pipeline optimization is turned off. With pipeline optimization, this dependency analysis process will be overlapped.) Therefore, the dependency analysis will not become a bottleneck.

\begin{table}[ht]  
\vspace{-0.10in}
\centering
    \captionsetup{font=small}
    \caption{Performance comparison  with dependency and without dependency}
    \label{tab:with_dep}
    \vspace{-0.1in}
    \footnotesize
    {\setlength{\tabcolsep}{0.9mm}{
    \begin{tabular}{l | r | r | r }
    \hline

    \hline
    \multirow{2}{*}{\textbf{Dataset}} &
    \multicolumn{2}{c|}{\textbf{Runtime of 80 epochs (s)}} & \multirow{2}{*}{\textbf{Speedup}}  \\ 
    \cline{2-3}
     & With Temporal Dependency\ & Without Temporal Dependency &  \\
     \hline
      DyRep & 8592.48 & 4544.05  & 1.89X \\ 
      \hline
      DGNN & 25521.12 & 16714.96 & 1.53X \\

    \hline

    \hline
    \end{tabular}}}
    \vspace{-0.1in}
\end{table}

\section{EXPERIMENTAL EVALUATION}

\subsection{Experimental Setup}
\Paragraph{Environments} Our experiments are conducted on an Aliyun ECS server (ecs.g5.6xlarge instance) equipped with 24 vCPU cores, 96GB DRAM. The server runs Ubuntu 18.04 LTS OS with GCC-7.5 as the host compiler. Libraries OpenMP-4.0.3, PyTorch v1.12.1 backend \cite{Pytorch_NIPS_2019} are used in the server.

\Paragraph{Models} 
We select three Dynamic GNN models: DyRep \cite{DyRep_ICLR_2019} , LDG \cite{LDG_Plosone_2021} and DGNN \cite{DyGNN_SIGIR_2020}. 
\addrevision{Most existing stream-based dynamic GNNs models can be divided into two categories: \textit{temporal point-based model} and \textit{RNN-based model}. DyRep and LDG are two representative temporal point-based models, while DGNN is a representative RNN-based model.}
In our experiments, we focused on the one-hop neighbors of event nodes because DyRep \cite{DyRep_ICLR_2019} and LDG \cite{LDG_Plosone_2021} aggregate the one-hop neighbor information through the attention mechanism, while DGNN \cite{DyGNN_SIGIR_2020} diffuses the updated event node embedding to one-hop neighbors. We set the dynamic node embedding size to be 64.

\begin{table}[t]  
\vspace{-0.1in}
    \captionsetup{font=small}
    \caption{\addrevision{Dataset statistics}}
    \vspace{-0.1in}
    \label{tab:dataset}
    \footnotesize
    \setlength{\tabcolsep}{3.1mm}{
    \begin{tabular}{l r r r r}
    \toprule
    Dataset & $|V|$ & $|E|$.init & $|E|$.final & $evt.num$ \\
    \midrule
    Social Evolution \cite{dataset_social_evolution} & 84    & 575       & 794          & 54,369    \\
    Github  \cite{github_dataset}        & 284   & 298       & 4,131         & 20,726  \\
    DNC \cite{dnc_uci_dataset_WWW_2013}            &2,029   & 0         & 5,598         & 39,264 \\
    UCI \cite{dnc_uci_dataset_WWW_2013}            &1,899   & 0         &20,296         & 59,835  \\
    \addrevision{Reality   \cite{Reality_Journal_2006}}      &\addrevision{6,809}  & \addrevision{0}         &\addrevision{9,484}          &\addrevision{52,052}      \\
    \addrevision{Slashdot  \cite{slashdot_WWW_2008}}       &\addrevision{51,083}  & \addrevision{0}         &\addrevision{131,175}     &\addrevision{140,778}     \\
    \bottomrule
    \end{tabular}}
\vspace{-0.15in}
\end{table}


\Paragraph{Datasets} Table \ref{tab:dataset} shows the six datasets used in our experiments.
Social Evolution \cite{dataset_social_evolution} and Github \cite{github_dataset} are two social network datasets. DNC \cite{dnc_uci_dataset_WWW_2013}, UCI \cite{dnc_uci_dataset_WWW_2013}, Reality   \cite{Reality_Journal_2006} and Slashdot \cite{slashdot_AAAI_2015} are communication network datasets.
In Social Evolution and Github, nodes represent users, and edges represent various type of interaction events. 
In DNC, UCI, Reality, and Slashdot, nodes represent users, and edges represent communication events.
For Social Evolution graph, we followed DyRep \cite{DyRep_ICLR_2019} and LDG \cite{LDG_Plosone_2021}, configuring \emph{CloseFriend} events as association events and other types of events as communication events. For Github, We use \emph{Follow} events as association events and other types of events as communication events. 

\Paragraph{Baselines}
We implement three training methods on \oursys: batch, sliding window, and adaptive sliding window. Batch training refers to accumulating a batch of new events from the event stream as the training set input to the model, denoted as \batch. We set the batch size to 200.
Sliding window refers to sliding a certain stride each time and selecting a fixed window event as a training set, denoted as \sliding (Algorithm \ref{alg:incremental_learning}). We set the stride to 40 and the window size to 200. The adaptive sliding window refers to adaptively adjusting the window size based on the sliding window method to capture continuous events with locality, denoted as \texttt{\adasliding}  (Algorithm \ref{alg:auto_window}). We set the range of a window from 100 to 300, with a stride of 20 \% of the 
window size.

In \batch, we take the 200 events adjacent to each batch as the test set and divide it into 5 test units. We feed a test unit into the model for testing after every 20 epochs of training. Subsequently, we feed these 200 events into the model as the next training batch.
In \sliding and \adasliding, we perform 20 training epochs on each window. In the sliding window, we select the 40 events adjacent to each window as the test set. Subsequently, we add these 40 events to the next window for training. 
In \adasliding, we select the last 20\% of events within a window as the test set. 
Based on the above settings, we ensure equal training iterations for each event in each method for fair comparison. 
We use link prediction as the downstream task and set the number of negative samples to 5. 
In performance experiments, we also compare the batch method based on the open-source implementations of the three models: \cite{dyrep_code}, \cite{ldg_code}, \cite{dgnn_code}.
These models are implemented based on PyTorch \cite{Pytorch_NIPS_2019}, so we denote them as \torchbatch.
\addrevision{\oursys also uses the C++ version LibTorch of PyTorch as the underlying operator library. Therefore, it is fair to compare \oursys with PyTorch. The open-source codes cannot be further optimized in PyTorch, because the key to affecting the training performance of dynamic GNNs is the sequential nature of time-dependent events. }

\vspace{-0.11in}
\subsection{Runtime Comparison}


\begin{figure*}[ht] 
\vspace{-0.35in}
\flushleft
\begin{minipage}{.78\textwidth}
    \begin{center}
    \includegraphics[width=1\linewidth]{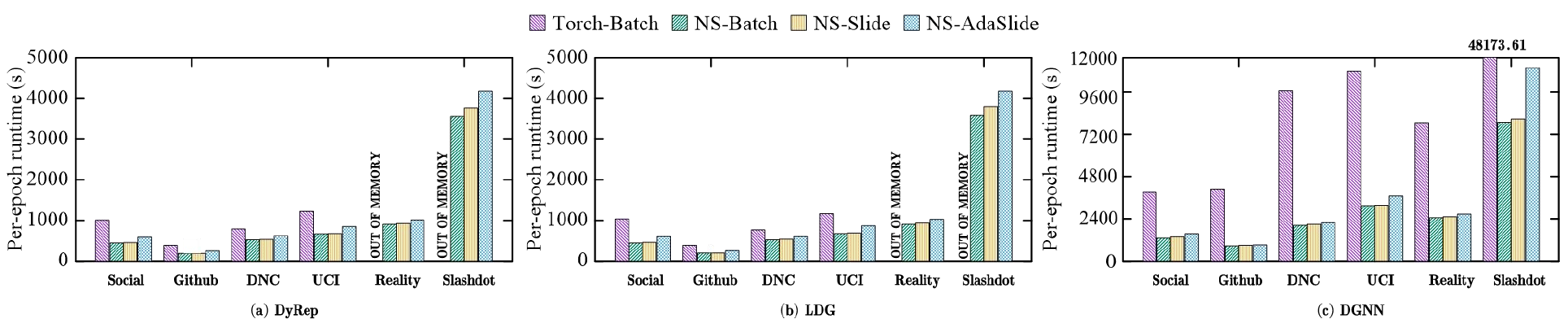}
      \vspace{-0.3in}
    \captionsetup{font=small}
    \caption{Runtime comparison.}
    \label{fig:overall}
    \end{center}
\end{minipage}
\begin{minipage}{.20\textwidth}
    \begin{center}
    \includegraphics[width=1\linewidth]{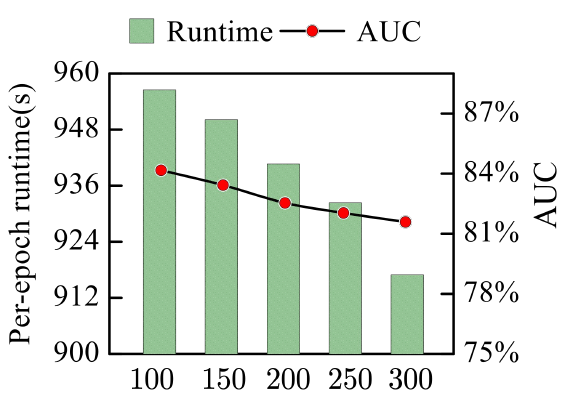}
      \vspace{-0.28in}
    \captionsetup{font=small}
    \caption{\addrevision{Varying window size in \sliding.}} \label{fig:varying_win_size}
     \vspace{-0.1in}
    \end{center}
\end{minipage}
 \vspace{-0.2in}
\end{figure*}


\begin{figure*}[ht] 
\vspace{0.05in}
\flushleft
\begin{minipage}{.35\textwidth}
    \begin{center}
    \includegraphics[width=0.85\linewidth]{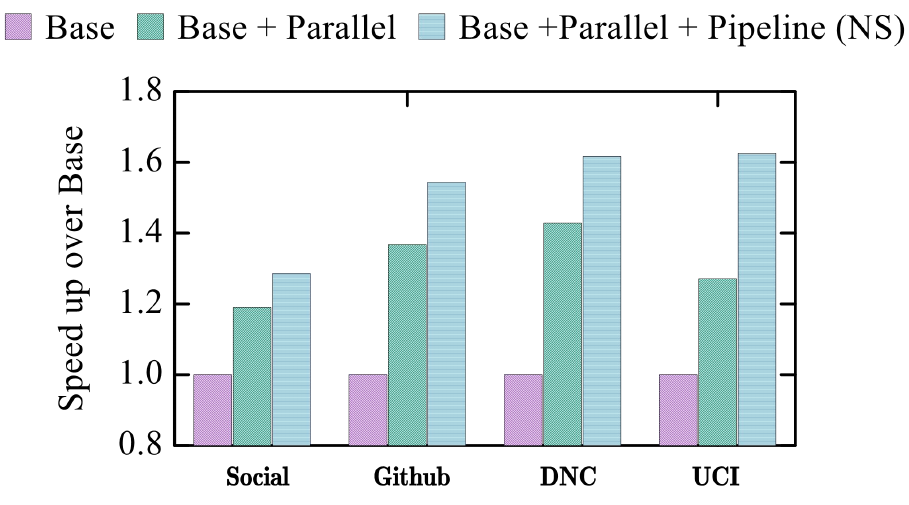}
      \vspace{-0.13in}
    \captionsetup{font=small}
    \caption{Performance breakdown\\ analysis.}
    \label{fig:gain}
    \end{center}
\end{minipage}
 \hspace{-0.15in}
\begin{minipage}{.20\textwidth}
    \begin{center}
    \includegraphics[width=1.1\linewidth]{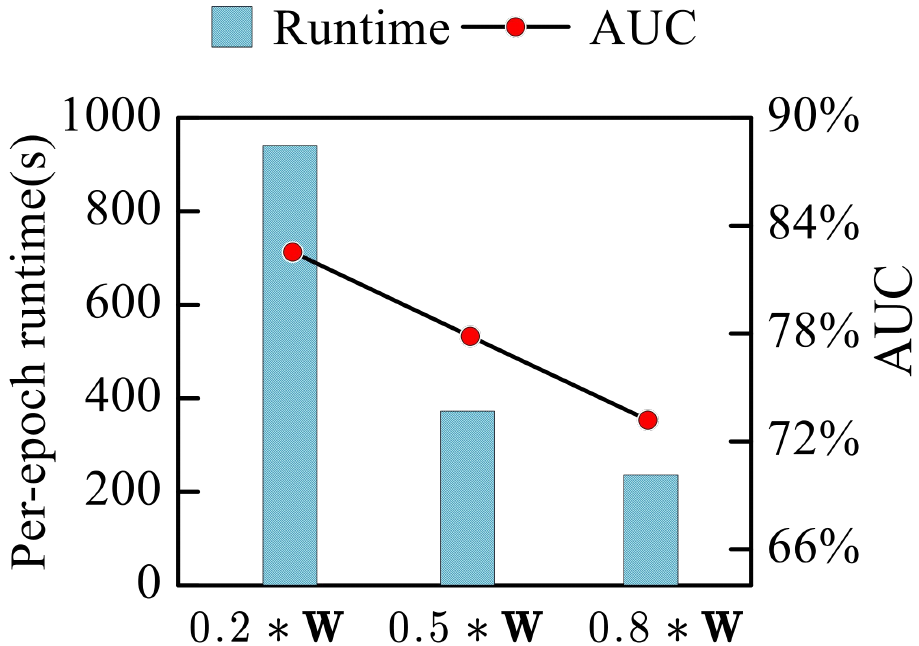}
      \vspace{-0.23in}
    \captionsetup{font=small}
    \caption{\addrevision{Varying sliding stride in \sliding.}} \label{fig:varying_slide_stride}
     \vspace{-0.025in}
    \end{center}
\end{minipage}
\hspace{0.1in}
\begin{minipage}{.30\textwidth}
    \begin{center}
    \includegraphics[width=1.26\linewidth]{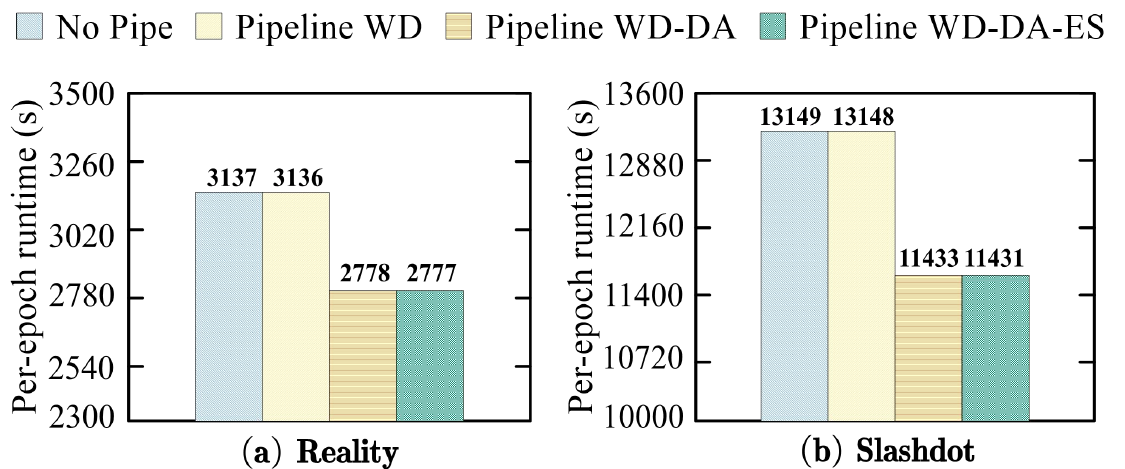}
     \vspace{-0.23in}
     \captionsetup{font=small}
    \caption{\addrevision{Pipeline ablation.}}\label{fig:pipe_ab}
     \vspace{0.12in}
    \end{center}
\end{minipage}

 \vspace{-0.1in}
\end{figure*}

We first compare the overall performance of the three training methods implemented based on \oursys and the batch method implemented based on open-source codes.  
Figure \ref{fig:overall} reports the runtime of training one epoch on different datasets. All the per-epoch runtime results are measured by averaging results of 10 epochs.


Compared to \torchbatch , \batch can achieve 1.53X-2.26X speedups over DyRep, 1.48X-2.33X speedups over LDG, and 3.01X-5.87X speedups over DGNN, demonstrating the effectiveness of our system optimization. LDG and DyRep share the same underlying code and storage structures, so their performances are similar. However, DGNN involves more complex computations and performs more read and write operations,  so the speedup of our system is more obvious. 
In \adasliding, the window captures the locality between events, resulting in lower parallelism, thus requiring a longer training time. 
However, thanks to our system optimization, \adasliding can also achieve speedup improvements compared to \torchbatch . \adasliding achieves 1.28X-1.71X speedups over DyRep, 1.27X-1.72X speedups over LDG, and 2.53X-4.44X speedups over DGNN.
DyRep and LDG use uncompressed matrices to store graph topology and node embeddings. They need to perform multiple memory allocation and copy operations to store the updated embeddings. Therefore, they report the ``Out-Of-Memory'' error on two large datasets ``Reality'' and ``Slashdot''. Compared with the open-source codes of DGNN, \batch achieves speedup 3.20X and 5.87X on these two datasets, respectively. Our framework can support training on larger datasets with more events, which demonstrates the scalability of our approach.

\vspace{-0.10in}
\subsection{AUC Comparison}
We use the Area Under the ROC Curve (AUC) metric on the test set to evaluate the performance of models. 
We sum the highest AUC for each batch or window and compute the average as a result, as shown in Table \ref{tab:auc}.
\torchbatch and \batch represent the batch training implemented based on PyTorch and our system, respectively. They use the same training method, so their AUC performance is identical. Among the three methods, the batch method has the lowest AUC because it cuts off the data stream, resulting in the loss of training information. 
The sliding window method reduces information loss and improves AUC compared with the batch method. The adaptive sliding method achieves the highest AUC by effectively capturing the spatial-temporal locality between events. 
The experimental results demonstrate the effectiveness of the sliding window method and the adaptive sliding window method.

\begin{table}[t]  
\centering
    \captionsetup{font=small}
    \caption{AUC comparison of three training methods}
    \vspace{-0.1in}
    \label{tab:auc}
    \footnotesize
    \setlength{\tabcolsep}{1.6mm}{
    \begin{tabular}{l | l | c | c | c | c }
    \hline

    \hline
    \multirow{2}{*}{\textbf{Model}} & \multirow{2}{*}{\textbf{Dataset}} &
    \multicolumn{4}{c}{\textbf{Training Method}}  \\ 
    \cline{3-6}
     &&\torchbatch & \batch & \sliding & \adasliding  \\
    
    \hline
    \multirow{4}*{DyRep} &  Social & 83.01$\%$ & 83.01$\%$ & 86.15$\%$ & \textbf{89.32$\%$} \\
     \cline{2-6}
     & Github & 73.46 $\%$ & 73.46$\%$ & 77.54$\%$ & \textbf{79.28$\%$} \\
     \cline{2-6}
     & DNC & 63.15$\%$ & 63.15$\%$ & 63.49$\%$ & \textbf{66.18$\%$} \\
     \cline{2-6}
     & UCI & 62.46$\%$ & 62.46$\%$ & 63.53$\%$ & \textbf{65.68$\%$} \\

    \hline
    \multirow{4}*{LDG} &  Social & 87.07$\%$ & 87.07$\%$ & 88.52$\%$ & \textbf{92.98$\%$} \\
    
     \cline{2-6}
     & Github & 74.34$\%$ & 74.34$\%$ & 78.10$\%$ & \textbf{79.16$\%$} \\
     \cline{2-6}
     & DNC & 64.62$\%$ &  64.62$\%$ & 66.50$\%$ & \textbf{69.41$\%$} \\
     \cline{2-6}
     & UCI &62.16$\%$ & 62.16$\%$ & 64.38 $\%$& \textbf{66.57$\%$} \\

    \hline
    \multirow{4}*{DGNN} &  Social & 97.21$\%$ & 97.21$\%$ & 97.61$\%$ & \textbf{97.67$\%$} \\
    
     \cline{2-6}
     & Github & 81.94$\%$ & 81.94$\%$ & 82.54$\%$ &  \textbf{84.45$\%$} \\
     \cline{2-6}
     & DNC & 86.04$\%$ &86.04 $\%$ & 87.72$\%$ & \textbf{88.81$\%$} \\
     \cline{2-6}
     & UCI & 78.45$\%$ & 78.45$\%$ &81.14 $\%$& \textbf{82.09$\%$} \\
    \hline

    \hline
    \end{tabular}}
    \vspace{-0.20in}
\end{table}

\vspace{-0.10in}
\subsection{Performance Gain Analysis}

We quantify the gains of event-parallel execution and pipeline optimization separately in the adaptive window method. We take the non-optimized version (\texttt{Base}) as the baseline and gradually integrate these two optimization methods into the baseline version. 
DyRep, LDG, and DGNN show similar trends, so we show the results of DyRep.
Figure \ref{fig:gain} shows the normalized speedup over the raw base processing.
Compared with \texttt{Base}, the parallel optimization (\texttt{Base+Parallel}) achieves 1.19X-1.43X speedups. On the Social dataset, the parallel speedup is lower because it is a small dataset with more dependencies between events. On the DNC dataset, the parallel speedup is higher because the dataset is sparse, and there are fewer dependencies between events. 
In addition, since our experiments are performed in the adaptive window method, this further limits parallel optimization. 
The pipeline optimization (\texttt{Base+Parallel +Pipeline}) can further achieve 1.08X-1.28X speedups compared to the parallel optimization. As the graph size increases, the time required for dependency analysis also increases, making the pipeline speedup more obvious. Therefore, the speedup from pipelining is more significant on the UCI dataset. Regarding the Social dataset, the speedup is relatively low due to its small size.

\begin{figure*}[t]
\vspace{-0.35in}
  \centering
  \includegraphics[width=6in]{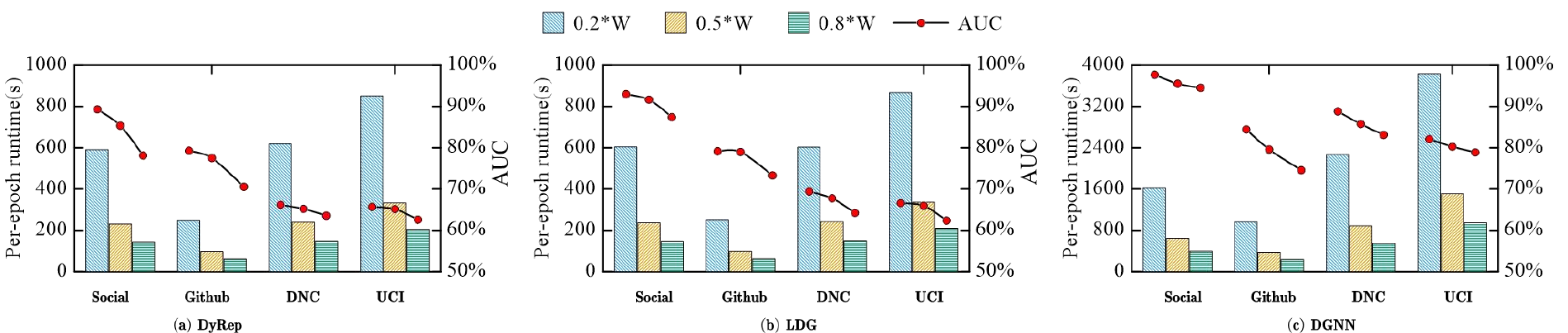}
  \vspace{-0.15in}
  \captionsetup{font=small}
  \caption{Runtime and AUC comparison when varying stride}
  \label{fig:varying_stride}
  \vspace{-0.2in}
\end{figure*}

\vspace{-0.23in}
\addrevision{\subsection{Evaluation the Impact on Parameters}}
\Paragraph{Varying Sliding Stride in \adasliding}
The stride determines the sliding stride of the window each time. 
We set the sliding stride as 0.2, 0.5, and 0.8 of the window size $W$ respectively to study the effect of sliding stride. As shown in Figure \ref{fig:varying_stride}, the smaller the sliding stride, the less information is lost in model training. Therefore, the model AUC is the highest when the stride is $0.2*W$, and the model AUC is the lowest when the stride is $0.8*W$. Although a smaller stride leads to higher model AUC, it also results in a longer training time. Our results show that setting the stride to $0.5*W$ achieves an average speedup of 2.55X compared to setting the stride to $0.2*W$. The stride of $0.8*W$ achieves an average speedup of 4.09X over the stride of $0.2*W$. Therefore, the sliding stride setting balance the training efficiency and the training accuracy. 

\Paragraph{Varying Sliding Stride in \sliding}
We set the sliding stride as 0.2, 0.5, and 0.8 of the window size $W$, where $W$ equals 200. \adasliding and \sliding show the same trends in accuracy and performance when varying sliding stride. The smaller the stride, the less information is lost and the accuracy is higher, but it also results in longer training time. As shown in Figure \ref{fig:varying_slide_stride}, the accuracy is the highest when the stride is 0.2, but it also takes the most training time. The accuracy is the lowest when the stride is 0.8, but it takes the shortest training time. In addition, the accuracy of varying stride is 84.45\%, 79.60\%, and 74.58\% in \adasliding, respectively. The accuracy of varying stride is 82.54\%, 77.85\%, and 73.19\% in \sliding, respectively. \adasliding achieves higher accuracy than \sliding when the ratio of stride is the same, which demonstrates the effectiveness of the adaptive sliding window method.

  

\vspace{-0.059in}
\Paragraph{Varying Window Size} 
We vary the window size between 100 and 300, and set the sliding stride to 0.2 of the window size. We show the AUC  and the per-epoch runtime of DGNN on Github, as shown in Figure \ref{fig:varying_win_size}. The accuracy increases when the window size decreases from 300 to 100. This is because Github has low spatial-temporal locality, and smaller windows can sufficiently capture the locality among events. Smaller windows also have a smaller stride, reducing information loss. In addition, when the window size varies from 100 to 300,  the sliding window method achieves the highest accuracy of 84.17\% when the window size is 100. The adaptive sliding window method can reach 84.45\%, which demonstrates the effectiveness of the adaptive method. In performance, different window sizes show similar running times as their training events are similar. In addition, as the window size increases, the parallelism increases, so the window size of 300 shows the shortest runtime.

\vspace{-0.03in}
\Paragraph{Varying Parameters $H$ and $L$} 
We set five ranges: [50, 300], [100, 300], [150, 300], [100, 250], [100, 350]. The left number indicates the setting of $L$, and the right number indicates the setting of $H$. We set the sliding stride to 0.2 of the window size. We show the AUC and the per-epoch runtime of DGNN on Github, as shown in Table \ref{tab:varying_win_parameters}. 
When fixing $H$ and varying $L$, a smaller $L$ leads to higher accuracy. This is because Github has low spatial-temporal locality, and smaller $L$ can better capture the locality among events.
However, when fixing $L$ and varying $H$, they show the same accuracy since each window does not contain more than 250 events.
Therefore, the parameter $L$ significantly impacts accuracy for datasets with low temporal-spatial locality.
In performance, different settings show similar runtimes as their training events are similar. In addition, as the parameter $L$ increases, the parallelism increases, so the setting with [150,300] shows the shortest runtime.

\vspace{-0.10in}
\begin{table}[ht]
        \captionsetup{font=small}
	\caption{\addrevision{Performance and accuracy when varying $L$ and $H$ (DGNN on Github)}}
        \vspace{-0.1in}
	\label{tab:varying_win_parameters}
	\footnotesize
	{\setlength{\tabcolsep}{1.5mm}{
	\begin{tabular}{l | r | r | r | r | r}
		\hline
  
		\hline
		\multirow{2}*{} &
		\multicolumn{5}{c}{\textbf{Varying Window Size Parameters $H$ and $L$ in \adasliding}}\\
		\cline{2-6}
		&{[50, 300]} &
		{[100, 300]}  &
		{[150, 300]}&
		{[100, 250]}&
		{[100, 350]}\\
		\hline
		{Accuracy ($\%$) }& 85.61 & 84.45 & 83.47 & 84.45 & 84.45 \\
		\hline
		{Peformance (s)}& 1015.13 & 963.02 & 950.15 & 966.59 & 959.68 \\
		\hline

		\hline
	\end{tabular}}
	}
    \vspace{-0.2in}
\end{table}

\begin{table}[ht]  
\centering
    \captionsetup{font=small}
    \caption{Graph query performance comparison}
    \vspace{-0.1in}
    \label{tab:query}
    \footnotesize
    \setlength{\tabcolsep}{3.7mm}{
    \begin{tabular}{l | l | c | c }
    \hline

    \hline
    \multirow{2}{*}{\textbf{Model}} & \multirow{2}{*}{\textbf{Dataset}} &
    \multicolumn{2}{c}{\textbf{Query Time (s)}}  \\ 
    \cline{3-4}
     && Torch\_Query & NeurtonStream\_Query \\
    
    \hline
    \multirow{4}*{DyRep} &  Social & 281.40 & 114.38  \\
     \cline{2-4}
     & Github & 101.76 & 47.29 \\
     \cline{2-4}
     & DNC & 143.78 & 101.62 \\
     \cline{2-4}
     & UCI & 223.04 & 141.03 \\

    \hline
    \multirow{4}*{LDG} &  Social & 278.46 & 114.85 \\
    
     \cline{2-4}
     & Github & 108.2 & 46.88 \\
     \cline{2-4}
     & DNC & 146.75 & 101.68 \\ 
     \cline{2-4}
     & UCI & 220.71 & 140.73 \\

    \hline
    \multirow{4}*{DGNN} &  Social & 762.69 & 329.30 \\
    
     \cline{2-4}
     & Github & 767.71 & 200.58 \\
     \cline{2-4}
     & DNC & 1968.62 & 419.90 \\
     \cline{2-4}
     & UCI & 2601.55 & 509.09 \\
    \hline

    \hline
    \end{tabular}}
    \vspace{-0.20in}
\end{table}

\vspace{0.2in}
\subsection{Performance of Dynamic Graph Storage}
We evaluate the graph storage performance of \oursys by comparing it with 
open-source codes in the batch training.
We regard reading and writing graph topology and node embeddings during training as graph queries. \batch maintains the same sequential processing events as the PyTorch-based to ensure fairness, denoted as \texttt{NeutronStream\_Query}. 
Table \ref{tab:query} shows the per-epoch query time. 
\oursys can achieve 1.42X-2.46X speedups over DyRep, 1.44X-2.42X speedups over LDG, and 2.32X-5.11X speedups over DGNN. 
LDG and DyRep have the same underlying storage, 
so \oursys demonstrates similar speedups over these two models. DyRep and LDG use uncompressed matrices to store graph topology and node embeddings. 
They are highly inefficient in storing updated embeddings, requiring multiple memory allocations and copy operations. When computing each event, they first need to create a new matrix and copy from the previous event's embedding matrix, resulting in redundant storage.
DGNN uses a row-based sparse matrix (\texttt{scipy.sparse.lil\_matrix}) to store graph topology.
During graph query operations, DGNN needs to first convert the \texttt{lil\_matrix} to a COO-format matrix and then perform query operations. This model maintains multiple embeddings for each node. During the computation of an event, this model needs to read and write the embeddings of the event nodes and their neighbors, resulting in complex read-and-write operations. 
In addition, these models all require determining the total number of nodes before training, so they do not support adding/removing nodes. 



\vspace{-0.10in}
\subsection{Analysis of Pipeline Benefits}

The training process of a window can be divided into four components: Window determination (WD), Dependency analysis (DA), Event scheduling (ES), and Parallel training (PT). We perform pipelined ablation experiments of DGNN on Reality and Slashdot. Figure \ref{fig:pipe_ab} reports the per-epoch runtime when adding each component sequentially to the pipeline. No Pipe indicates four components are executed sequentially. Pipeline WD indicates the window determination component is added to the pipeline. Pipeline WD-EA indicates both the window determination and the dependency analysis are added to the pipeline. Pipeline WD-DA-ES indicates the window determination, the dependency analysis, and the event scheduling are added to the pipeline. No Pipe is the slowest one due to its totally sequential execution. Pipeline WD does not reduce the runtime obviously since WD is fast. DA is a costly step, so adding DA to the pipeline can significantly reduce the runtime. Similar to WD, ES is also a lightweight step, so adding ES to the pipeline only slightly reduces the runtime. The results show that the benefits of pipeline optimization mainly come from overlapping DA. 

\vspace{-0.10in}
\section{RELATED WORK}
\Paragraph{Static GNN Frameworks} Static GNNs are a class of deep learning models to learn from static graphs \cite{GGSNN_ICLR_2016,HGNN_CoRR_2019, CNNonGraph_NIPS_2016, ESGCN_EMNLP_2017, 
GCN_ICLR_2017, GIN_ICLR_2019, gnn_survey_arxiv2019, GAT_ICLR_2018, GNNRS_CoRR_2020, Lasagne_IEEE_2021}. Many frameworks are designed specifically for static GNNs \cite{PyG_arXiv_2019, DGL_arXiv_2019, ALIGRAPH_VLDB_2019, ROC_MLSYS_2020, Seastar_EuroSys_2021, P3_OSDI_2021, GraphANGEL_2021, HET_2021, GNNLab_EuroSys_2022, Ginex_vldb_2022, ByteGNN_VLDB_2022, SANCUS_VLDB_2022, FOS_ICDM_2022, 
PCG_VLDB_2022, Sgs_2022, Pasca_2022, IOGraphSystem_2022,ducati}.
PyG \cite{PyG_arXiv_2019} is a GNN library based on PyTorch, which provides a general message-passing interface for users. 
DGL \cite{DGL_arXiv_2019} provides the graph as the 
programming abstraction. It provides users with flexible APIs to implement arbitrary message-passing computation. It also provides a distributed framework DistDGL \cite{DISTDGL_ARXIV_2020}.
Sancus \cite{SANCUS_VLDB_2022} is a decentralized full-graph GNN training framework. It introduces history embedding and actively creates asynchrony, which avoids a lot of communication and speeds up training.
PASCA \cite{Pasca_2022} proposes a new Scalable Graph Neural Architecture Paradigm. 
It implements an automated search engine to systematically explore high-performance and scalable GNN structures.
NeutronStar \cite{NeutronStar_2022_SigMod} proposes a hybrid dependency management approach 
which can adaptively select an appropriate dependency strategy for different dependent neighbors to accelerate distributed GNN training. Although these frameworks can effectively support static GNN training, their underlying storage and execution engines cannot effectively support dynamic graph updates and training of dynamic GNNs.

\vspace{-0.02in}
\Paragraph{Dynamic GNN Frameworks} Dynamic graphs represent graphs evolving over time. They can be categorized into snapshot-based dynamic graphs and continuous-time dynamic graphs based on discrete or continuous timestamps \cite{RLDGSurvey_2020, Encode_Decoder_Survey_CoRR_2022}. The snapshot-based dynamic graph is a list of snapshots, each of which keeps the graph state at a certain moment. For a graph whose nodes or edges are updated frequently, it is more efficient to use this method for storage. There is a class of dynamic GNN models specifically designed for the snapshot-based dynamic graph, such as EvolveGCN \cite{EvolveGCN}, DySAT \cite{DySAT_2020_WSDM}, and DynGEM \cite{DynGEM}. The continuous-time dynamic graph can be viewed as a stream of graph update events. This representation can capture all temporal information compared with the snapshot-based dynamic graph. There is a class of dynamic GNN models designed for the continuous-time dynamic graph, such as DyRep \cite{DyRep_ICLR_2019}, DGNN \cite{DyGNN_SIGIR_2020}, TGAT \cite{TGAT_2020_arXiv}, Zebra \cite{Zebra_vldb_2023} and TE-DyGE \cite{TE_DyGE_DASFAA_2023}. There are a limited number of frameworks available for dynamic GNNs. PyGT \cite{PyGT_CIKM_2021}, \addrevision{DynaGraph \cite{DynaGraph_sigmode_2022}} and PiPAD \cite{PiPAD_2023} are specifically designed for training on snapshot-based dynamic graphs. \addrevision{ Cambricon-G \cite{CambricoG_IEEEtrans_2022} designs a specialized accelerator to optimize various variant models of GCN \cite{GCN_ICLR_2017}, such as GraphSage \cite{GraphSage_NIPS_2017}, DiffPool \cite{Diffpool_2018}, DGMG \cite{DGMG_2018}, and EdgeConv \cite{EdgeConv_2019}. }
They do not support training models designed for graph streams. TGL \cite{TGL_VLDB_2022} is a temporal GNN training framework. It introduces Temporal-CSR, a data structure that stores temporal graphs and supports parallel sampling. This structure requires storing the entire dynamic graph statically before training. To speed up the training, TGL proposes an intra-batch parallel training method, disregarding the temporal dependencies between events within a batch.

\vspace{-0.10in}
\section{Conclusions and Future Work}
\label{sec:conclusion }


We present \oursys, a dynamic GNN training framework. The effectiveness of \oursys and its high performance are contributed by several components, including a sliding-window-based method to incrementally train models for capturing the spatial-temporal dependency of events, a fine-grained event parallel processing scheme and a series of system optimizations. We evaluate \oursys on three dynamic GNN models, DyRep, LDG, and DGNN. The experimental results demonstrate that our optimized sliding-window-based training brings 3.97\% accuracy improvements on average. Compared with their open-sourced implementations with PyTorch, \oursys achieves 2.26X-12.35X speedups. The learning models for graph streams are more desirable in real-world application scenarios. We believe that the online training support for GNNs is promising and can boost the wide applications of GNNs. Our future work will be tackling the challenges of online training for dynamic graphs.


\vspace{-0.10in}
\begin{acks}
 This work is supported by the National Natural Science Foundation of China (U2241212, 62072082, 62202088, and 62072083), the 111 Project (B16009), the Fundamental Research Funds for the Central Universities (N2216015 and N2216012), and Huawei. Yanfeng Zhang and Qiange Wang are the corresponding authors.
\end{acks}

\balance

\bibliographystyle{ACM-Reference-Format}
\bibliography{sample}

\end{document}